\documentclass[10pt,journal,compsoc]{IEEEtran}
\usepackage{amsmath,amsfonts}
\usepackage{algorithmic}
\usepackage{algorithm}
\usepackage{array}
\usepackage{textcomp}
\usepackage{stfloats}
\usepackage{url}
\usepackage{verbatim}
\usepackage{graphicx}

\ifCLASSOPTIONcompsoc
  \usepackage[nocompress]{cite}
\else
  \usepackage{cite}
\fi{\tiny {\tiny }}

\newcommand{\comm}[1]{\iffalse #1 \fi}
\usepackage{colortbl}
\usepackage{soul}
\usepackage{pifont}
\usepackage[table,dvipsnames,svgnames]{xcolor}
\usepackage{hyperref} 
\usepackage{lipsum}
\usepackage{hhline}
\usepackage{booktabs}
\usepackage{subcaption}
\usepackage{multirow}
\usepackage{multicol}
\usepackage{xspace}
\usepackage{balance}
\usepackage{makecell}
\definecolor{airforceblue}{rgb}{0.36, 0.54, 0.66}
\definecolor{revcolor}{rgb}{0.0, 0.0, 0.8}

\newcommand{\xmark}{\ding{55}}%
\newcommand{\etal}{~\textit{et~al.}\xspace}
\definecolor{Gray}{gray}{0.9}

\newif\ifhighlight
\highlightfalse
\ifhighlight
    \newcommand{\rev}[1]{\textcolor{revcolor}{#1}}
\else
    \newcommand{\rev}[1]{#1} 
\fi

\begin{document}

\title{A Survey on Deep Learning Techniques for Action Anticipation}

\author{Zeyun Zhong, Manuel Martin, Michael Voit, Juergen Gall, Jürgen Beyerer

\thanks{Zeyun Zhong and Jürgen Beyerer are with the Karlsruhe Institute of Technology (KIT) and also with Fraunhofer IOSB, Germany. Manuel Martin and Michael Voit are with Fraunhofer IOSB, Germany. Juergen Gall is with the University of Bonn and the Lamarr Institute for Machine Learning and Artificial Intelligence, Germany.}
}

\markboth{Journal of \LaTeX\ Class Files,~Vol.~14, No.~8, August~2015}%
{Shell \MakeLowercase{\textit{et al.}}: Bare Demo of IEEEtran.cls for Computer Society Journals}

\IEEEtitleabstractindextext{
\begin{abstract}
The ability to anticipate possible future human actions is essential for a wide range of applications, including autonomous driving and human-robot interaction. Consequently, numerous methods have been introduced for action anticipation in recent years, with deep learning-based approaches being particularly popular. In this work, we review the recent advances of action anticipation algorithms with a particular focus on daily-living scenarios. Additionally, we classify these methods according to their primary contributions and summarize them in tabular form, allowing readers to grasp the details at a glance. Furthermore, we delve into the common evaluation metrics and datasets used for action anticipation and provide future directions with systematical discussions.
\end{abstract}

}

\maketitle

\section{Introduction}
\begin{figure}[t]
  \includegraphics[width=\linewidth]{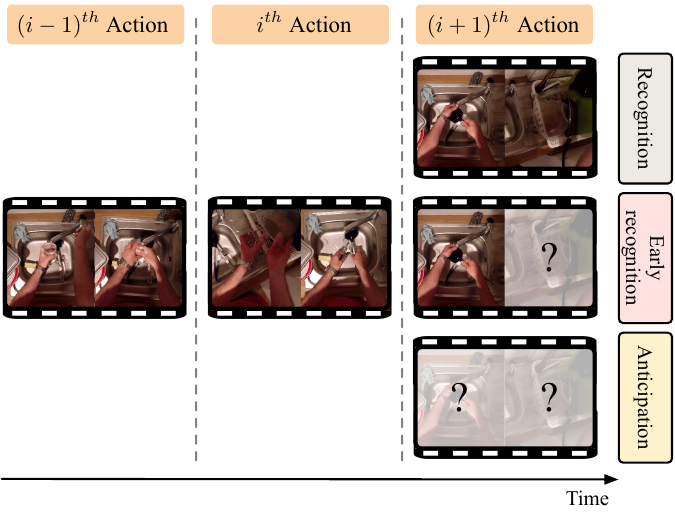}
  \vspace{-0.8cm}\caption{The action anticipation task aims to anticipate future actions before they happen, whereas action recognition and early action recognition require the observation of complete and partial actions, respectively. The example frames are from the EpicKitchens~\cite{damen2018scaling} dataset.}\vspace{-2mm}
  \label{fig:intro}
\end{figure}

Compared to human action recognition and early action recognition, where the entire or part of an action is observable, action anticipation aims to predict a future action without observing any part of it, as displayed in Figure~\ref{fig:intro}. Anticipating possible future daily-living actions is one of the most important tasks for human machine cooperation and robotic assistance, e.g., to offer a hand at the right time or to generate a proactive dialog to provide more natural interactions. 

\begin{figure}[t]
    \centering
    \includegraphics[width=0.95\linewidth]{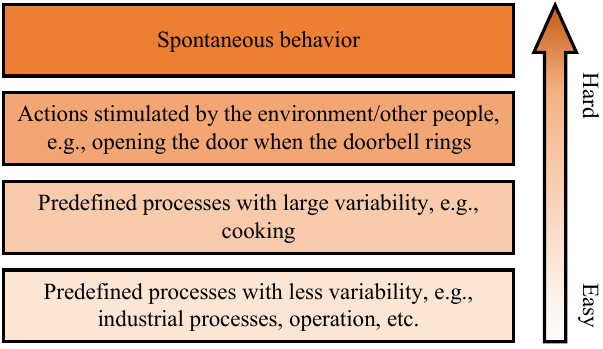}
    \caption{Predictability of actions.}
    \label{fig:predictability}
\end{figure}

As the future actions are often not deterministic, this tends to be significantly more challenging than the traditional action recognition task, where today’s well-honed discriminative models~\cite{feichtenhofer2019slowfast,liu2022video} perform very well.
The predictability of actions varies based on the nature of the tasks involved, as shown in Fig.~\ref{fig:predictability}. In the realm of predefined processes with less variability, such as industrial processes and medical operations~\cite{philipp2023formalisierung}, actions tend to be highly predictable due to the strict adherence to established protocols and guidelines. On the other hand, predefined processes with large variability, like cooking, introduce a degree of unpredictability as they involve subjective decision-making and varying ingredients, which can influence the final outcome. Actions stimulated by the environment or other people, exemplified by opening the door when the doorbell rings, exhibit moderate predictability, as they are influenced by external cues, but the responses are often ingrained through habitual patterns. Lastly, spontaneous behavior represents the highest level of unpredictability, as it arises without explicit stimuli or established patterns, making it challenging to anticipate. 

Recently, the computer vision research community has shown significant interest in addressing this challenging task.
In line with action recognition, anticipation approaches began with predictions based on a single video frame~\cite{vondrick2016anticipating} and tended to use longer temporal context in more recent works~\cite{senerTemporalAggregateRepresentations2020,wuMeMViTMemoryAugmentedMultiscale2022}. In addition to utilizing a longer action history, many approaches are exploring the incorporation of multiple modalities beyond the raw RGB video frames to further improve the anticipatory ability. These include optical flow, which contains motion information, as well as objects present in the scene, among others.

Apart from the aforementioned action recognition and early action recognition task, there are several other tasks in the literature that are related to action anticipation, such as action segmentation~\cite{ding2022surveyactionsegmentation}, temporal/spatio-temporal action detection~\cite{vahdani2022actiondetection}, video prediction~\cite{opreaReviewDeepLearning2020}, trajectory prediction~\cite{rudenko2020trajectory}, and motion prediction~\cite{lyu2022motion}. The definitions of action anticipation and its related tasks in computer vision are listed in Table~\ref{tab:related_tasks}. In contrast to action anticipation, the related tasks are concerned with recognizing and understanding actions that have already occurred or are in progress. Despite these differences, these tasks complement each other, with action recognition and localization providing foundational components for action anticipation.

\begin{table}[t]
    \centering
    \caption{Definition of action anticipation and related tasks.}
    \begin{tabular}{@{}p{2.5cm}p{5.9cm}@{}}
        \toprule
        Task & Definition \\
         \midrule
         Action Anticipation & Anticipates one or multiple future actions before they happen.\\ \midrule
         Action Recognition & Categorizes actions in a video or image sequence given the full observation. \\ \midrule
         Early Action \newline Recognition & Categorizes actions in a video or image sequence given the partial observation. \\ \midrule
         Action \newline Segmentation & Categorizes actions for every frame of the video.\\ \midrule
         Temporal Action \newline Detection & Identifies and localizes actions within a video by predicting the start and end times of actions. \\ \midrule
         Spatio-temporal \newline Action Detection & Identifies and localizes actions in both spatial and temporal dimensions within a video. \\ \midrule
         Video Prediction & Predicts future frames given a set of video frames.\\ \midrule
         Trajectory Prediction & Estimates the future trajectory or path of objects or agents in a scene. \\ \midrule
         Motion Prediction & Predicts changes in human pose.\\
         \bottomrule
    \end{tabular}
    \label{tab:related_tasks}
\end{table}

\subsection{Application Domains}
While exploring the plausible future is well studied in other fields, such as weather forecasting~\cite{shi2015convolutional} and stock price prediction~\cite{mackie2004price}, researchers have only recently become heavily active in exploring solutions in the computer vision community. 
As such, action anticipation approaches have been applied for diverse applications across various domains in recent years, such as industrial applications~\cite{petkovicHumanIntentionEstimation2019}, robotics~\cite{koppulaAnticipatingHumanActivities2016}, advanced driver assistance systems~\cite{jainCarThatKnows2015}, autonomous driving~\cite{rasouliPedestrianActionAnticipation2019}, and more. 

In the industrial domain like robotized warehouses, action anticipation approaches can be used to infer the intention of workers to improve efficiency and safety~\cite{petkovicHumanIntentionEstimation2019,alati2019help}. 
In the field of human-robot-interaction, in contrast to passive service, e.g., providing service only after being spoken to, robots are expected to provide proactive service that initiates an interaction at an early stage. In this context, the ability of accurately predicting the start of an interaction~\cite{itoAnticipatingStartUser2020} or the object that the user is going to interact with~\cite{koppulaAnticipatingHumanActivities2016,schydlo2018anticipation,huang2015using} enables robots to respond naturally and intuitively to human actions, improving collaboration. 
Anticipation can also find application in generating notifications for our daily routines, such as reminding us to turn off lights before leaving a room or the stove after cooking~\cite{soran2015notification}.
Furthermore, the safety of users, particularly those in need of care,
can be significantly improved if robots are able to assess the risks
of their current environment~\cite{zengAgentCentricRiskAssessment2017}.
In driving scenarios, anticipating a dangerous maneuver beforehand can alert drivers before they perform the maneuver and can also give driver assistance systems more time to avoid or prepare for the danger~\cite{jainCarThatKnows2015}.
In autonomous driving, intelligent vehicles rely on predictive abilities to anticipate the actions of other road users in urban environments, particularly pedestrians at crosswalks~\cite{rasouliPedestrianActionAnticipation2019} or potential traffic accidents~\cite{suzuki2018trafficaccidents}, ensuring safe and efficient navigation.

\subsection{Review Scope and Terminology}
There are several surveys in the literature that discuss action anticipation as part of their scope~\cite{trong2017surveyactivityprediction,rasouli2020deep,kong2022survey,rodin2021predicting,hu2022online,plizzari2023outlook}. 
However, these surveys do not provide a full overview of the action anticipation literature.  
While \cite{rasouli2020deep,kong2022survey,hu2022online} consider several topics and action anticipation is only briefly discussed among other topics, \cite{rodin2021predicting,plizzari2023outlook} focus on the specific challenges of egocentric vision. \cite{trong2017surveyactivityprediction}, being one of the earlier works, primarily delves into methods that are not deep learning-based.
In this survey, our objective is to review and classify the research field of action anticipation in general. \comm{However, we will make an exception for automotive scenarios due to the significant divergence in input cues between automotive and daily-living scenarios.}
This includes an overview of the current state-of-the-art approaches as well as datasets and evaluation metrics. We conclude our review by identifying future research challenges and by summarizing our findings.

Although classical learning approaches, such as models based on the Markovian assumption, including Probabilistic Suffix Tree (PST)~\cite{liModelingComplexTemporal2012,liPredictionHumanActivity2014}, Hidden Markov Model (HMM)~\cite{jainCarThatKnows2015}, Conditional Random Fields (CRFs)~\cite{koppulaAnticipatingHumanActivities2016}, Markov Decision Process (MDP)~\cite{rhinehart2017first}, and other statistical methods~\cite{mahmud2016poisson,qiPredictingHumanActivities2017}, have been widely used in the literature, we put our focus on deep learning techniques and how they have been extended or applied to action anticipation, leaving the classical approaches outside the scope of the present review. In this context, the terms action anticipation, action prediction, and action forecasting are used interchangeably. Note that we do not explicitly differentiate between the terms action and activity in this survey, since these two terms are commonly used interchangeably in the literature.

\section{Problem Statement}
\label{sec:problem_statement}

Based on the predicted time horizon, action anticipation approaches can be grouped into two categories: short-term anticipation approaches and long-term anticipation approaches. Short-term approaches usually operate on subsymbolic sensory data and predict actions a few seconds into the future. Conversely, long-term approaches may utilize action history with a higher abstraction level to predict a sequence of future actions (with their durations) up to several minutes into the future. 
In Section~\ref{sec:feature_encoding}, we first introduce video feature encoding, a foundational step for action anticipation, and then present the detailed task definitions commonly used in the literature for both categories in Section~\ref{sec:short-term_task} and Section~\ref{sec:long-term_task}, respectively.

\subsection{Video feature encoding}
\label{sec:feature_encoding}
\rev{
Compared to \textit{trimmed} videos that are clipped to isolate a single action~\cite{carreiraQuoVadisAction2017}},
\textit{untrimmed} videos are typically used to evaluate action anticipation approaches. They can be as long as several minutes, and contain several actions. It is therefore difficult to directly input the entire video to a visual encoder for feature extraction due to the limits of computational resources. A common strategy for video representation is to partition the video into equally sized temporal intervals called snippets, and then apply a pre-trained feature extractor over each snippet. 

More formally, let the observed video segment be denoted by $V$ that contains $T$ frames $\{ x_1, x_2, \dotsc, x_T \}$, corresponding to $i$ actions $\{ a^1, a^2, \dotsc, a^i\}$, where $i$ is usually much smaller than $T$. Then, the video segment is broken down into $N$ snippets \{$V_1, V_2, \dotsc, V_N$\}, with each snippet $V_k$ containing $n$ video frames $\{ x_{k1}, x_{k2}, \dotsc, x_{kn}\}$, where the exact number of frames $n$ depends on the frame rate of the videos and the used feature extractor. Afterwards, a feature extractor is applied on these snippets to extract a sequence of video representations \{$z_1, z_2, \dotsc, z_N$\}.

Common extractors range from frame-level spatial models, such as VGG16~\cite{simonyan2014vgg}, ResNet50~\cite{he2016deep}, TSN~\cite{wangTemporalSegmentNetworks2016} and ViT~\cite{dosovitskiyImageWorth16x162021}, to snippet-level spatiotemporal models, such as I3D~\cite{carreiraQuoVadisAction2017}, Two-stream~\cite{simonyan2014twostream}, R(2+1)D~\cite{tran2018closer}, MViT~\cite{li2022mvitv2}, and Swin~\cite{liu2022video}. 
The features extracted from pre-trained encoders, which are typically trained for trimmed action recognition tasks, are not necessarily suitable for action anticipation.

\subsection{Short-term anticipation}
\label{sec:short-term_task}

\begin{figure}[t]
    \centering
    \includegraphics[width=\linewidth]{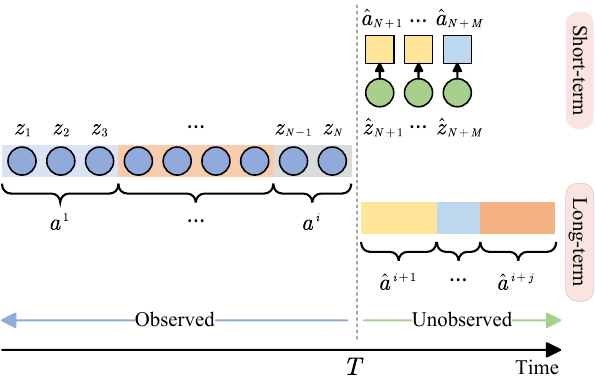}\vspace{-2mm}
    \caption{Category of the action anticipation task. Short-term anticipation aims to predict actions at several future time steps, whereas the long-term task aims to predict subsequent actions (along with their durations) without adhering to fixed time steps.}\vspace{-2mm}
    \label{fig:problem_statement}
\end{figure}

Most short-term anticipation approaches follow the setup defined in~\cite{vondrick2016anticipating,gaoREDReinforcedEncoderDecoder2017,damen2018scaling}. Approaches tackling the short-term anticipation task typically operate on \comm{\textit{synchronous} input, i.e., }the sequence of video representations extracted by a pre-trained feature extractor, as introduced in Section~\ref{sec:feature_encoding}. Taking these representations with the same temporal spacing as input, short-term approaches predict a few representations $M$ time steps into the future $\{ \hat{z}_{N+1}, \hat{z}_{N+2}, \dotsc, \hat{z}_{N+M}\}$, which are then classified into actions $\{  \hat{a}_{N+1}, \hat{a}_{N+2}, \dotsc, \hat{a}_{N+M} \}$, as illustrated in Figure~\ref{fig:problem_statement}. The number of time steps into the future $M$ depends on the anticipation protocol and the temporal spacing of the video representations. For instance, $M$ would be 4 if the temporal spacing were set at 0.25 seconds and the chosen protocol aimed to predict an action in 1 second. 

\subsection{Long-term anticipation}
\label{sec:long-term_task}

Parallel research addresses the long-term anticipation task~\cite{farhaWhenWillYou2018,graumanEgo4DWorld0002022}. The goal is to anticipate the category (and the duration) of future actions for a given time horizon, which can extend up to several minutes, as illustrated in Fig.~\ref{fig:problem_statement}. In contrast to short-term approaches, long-term approaches are not confined to using subsymbolic video representations as their input. Certain methods leverage past actions, represented as $\{ a^1, a^2, \dotsc, a^i\}$, to directly forecast a sequence of future actions, $\{ \hat{a}^{i+1}, \hat{a}^{i+2}, \dotsc, \hat{a}^{i+j}\}$~\cite{farhaWhenWillYou2018,abufarhaUncertaintyAwareAnticipationActivities2019,zhao2020diverse}. These past actions could either be based on actual observations or derived from an action segmentation technique~\cite{richard2017weakly,farha2019ms-tcn}.
Approaches that tackle long-term anticipation typically rely on fully-labeled data, i.e., sequences labeled with all future actions and their durations. Notably, some methods operate within a weakly supervised setting, using only a few fully labeled sequences and primarily sequences where only the next action is labeled~\cite{zhang2021weakly}.

\section{Methods}
\begin{figure*}[t]
  \centering
  \includegraphics[width=0.97\linewidth]{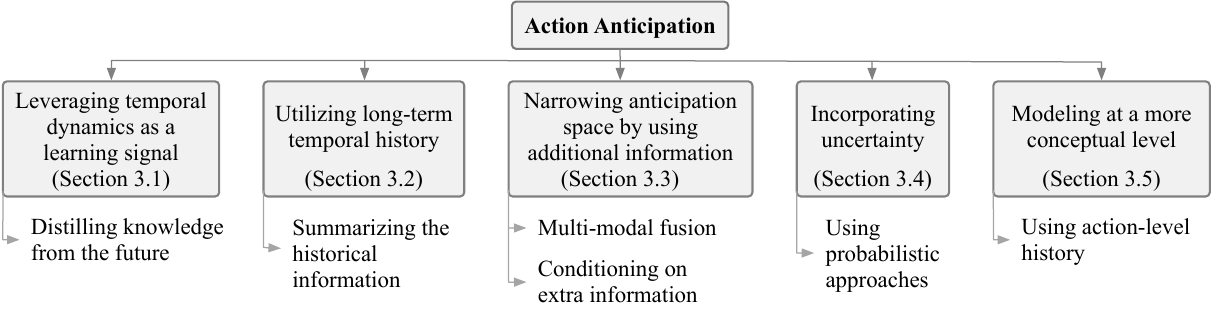}\vspace{-2mm}
  \caption{Classification of action anticipation methods.}\vspace{-2mm}
  \label{fig:classification_methods}
\end{figure*}

Although the predictive capability of machines could enable many real-world applications in robotics and manufacturing, developing an algorithm to anticipate the future action of the user is challenging. Humans rely on extensive knowledge accumulated over their lifetime to infer what will happen next. However, how machines can gain such knowledge remains an open question. 
In this section, we describe the methods that are trying to tackle such a challenging task. To our knowledge, the current literature does not provide a specific taxonomy to classify action anticipation models. In this review, we therefore classify the existing methods into five groups, according to the specific problem they addressed, as shown in Fig.~\ref{fig:classification_methods}. 

Leveraging the inherent temporal dynamics of videos as a learning signal, some methods build self-supervised frameworks to train the anticipation models (Section~\ref{sec:learning_signal}). Early methods anticipate future actions based on the representation of a single past frame, ignoring all temporal dynamics. To accurately predict the actions in the future, the evolution of the past actions should be analyzed and summarized. Consequently, recent approaches use various architectures to encode the long-term history, including recurrent and transformer-based models (Section~\ref{sec:history}). Some methods attempt to derive and utilize additional modalities\footnote{We refer to both different sensory data (e.g., RGB and audio) and different feature representations (e.g., RGB and optical flow) as \textit{modalities} for the sake of emphasizing the difference in data representations.} to improve the anticipative performance, such as presence of objects in the scene, object bounding boxes, and optical flow, as they may complement the raw RGB video frames. 
In addition to the data acquired from the input video, other information sources, e.g., personalization, the topology of the action sequences, or the underlying intention, may also contribute to improved future prediction (Section~\ref{sec:narrow}). As the vast majority of action anticipation models are deterministic, they do not deal with the uncertainty of the future. 
To address this issue, several authors proposed modeling uncertainty for future prediction (Section~\ref{sec:uncertainty}). Most action anticipation methods in the literature operate on the representation-level, i.e., future video representations are first anticipated and then categorized to actions, as illustrated in the upper part of Fig.~\ref{fig:problem_statement}. In contrast, some methods predict future actions directly based on the observed action sequences (Section~\ref{sec:concept}), which may provide a clean analysis of action dependencies and better explainability.

Table~\ref{tab:summary_models} shows an overview of the most relevant methods, ordered in chronological order. We characterize these approaches by their architecture, datasets used, and technical details, including modalities fed into the individual model, the fusion strategy (if any), incorporation of external knowledge, the abstraction level on which a prediction is made, anticipation horizon, and whether training is end-to-end or feature-based. 
Note that the taxonomy is not mutually exclusive, as some methods can be classified into several categories since they address multiple goals. For instance, \cite{abufarhaUncertaintyAwareAnticipationActivities2019,mehrasaVariationalAutoEncoderModel2019} address both action-level prediction and prediction uncertainty. We specify the category of these models according to their main contribution.

\begin{table*}[ht!]
    \centering
    \renewcommand*{\arraystretch}{1}
    \caption{Summary of most relevant action anticipation models. The following abbreviations are used in the table. Architecture: Transformer (TF), Non-local Block (NLB). Modalities: Objects (O), Bounding Boxes (BB), Motion (M), Action Classes (A), Audio (Au), Gaze (G), Hand Mask (H), Human-Object-Interaction (HOI), Time (T). Fusion: Score Fusion (score), Late feature fusion (late), Mid-level feature fusion (mid). EK: external knowledge. LT: Long-term anticipation. E2E: End-to-end. Because of space constraints, we have omitted the inclusion of object detectors such as Fast R-CNN~\cite{ren2015faster} and Mask R-CNN~\cite{he2017mask} as feature extractors for approaches that rely on object or human-object interaction modalities.}
    \setlength\tabcolsep{4pt}
    \resizebox{\linewidth}{!}{
    \begin{tabular}{@{}lcllllccccc@{}}
        \toprule
        \multirow{2}{*}{Method} & \multirow{2}{*}{Year} & \multirow{2}{*}{\shortstack[l]{Feature \\ extractor}} & \multirow{2}{*}{\shortstack[l]{Future\\ predictor}} & \multirow{2}{*}{Datasets} & \multicolumn{5}{c}{Details} \\ 
        \arrayrulecolor{gray} \cmidrule(lr){6-11} \arrayrulecolor{black}
        & & & & & Modalities & Fusion & EK & Abst. & LT &E2E \\ 
        \midrule
        
        \multicolumn{11}{c}{\textbf{Leveraging temporal dynamics as a learning signal}} \\
        \midrule
        Vondrick\etal \cite{vondrick2016anticipating} & 2016 & AlexNet\cite{Alexnet2012} & MLP & \cite{patron2010high,pirsiavash2012detecting,thumos2014}& RGB & -- & \xmark & feat. & \xmark & \checkmark\\
        Zeng\etal~\cite{zeng2017visual} & 2017 & ResNet\cite{he2016deep} & MLP & \cite{patron2010high,thumos2014} & RGB & -- & \xmark & feat. & \xmark & \checkmark\\
        RED~\cite{gaoREDReinforcedEncoderDecoder2017}& 2017 & VGG\cite{simonyan2014vgg},TS\cite{xiong2016twostream},& LSTM & \cite{geest2016online,patron2010high,thumos2014} & RGB,M & mid & \xmark & feat. & \xmark & \xmark\\
        Zhong\etal~\cite{zhong2018unsupervised}& 2018 & ResNet~\cite{he2016deep} & LSTM & \cite{patron2010high,thumos2014} & RGB & -- & \xmark & feat. & \xmark & \checkmark\\
        Tran\etal~\cite{tran2021knowledge} & 2021 & I3D~\cite{carreiraQuoVadisAction2017} & -- & \cite{damen2018scaling} & RGB & -- & \xmark & feat. & \xmark & \xmark \\
        Fernando\etal~\cite{fernandoAnticipatingHumanActions2021}& 2021 & R(2+1)D~\cite{tran2018closer}\comm{,F. r-cnn~\cite{ren2015faster}} & FC  & \cite{damen2018scaling,kuehneLanguageActionsRecovering2014} & RGB,O,M & score & \xmark & feat.,act. & \xmark & \checkmark\\

        \rev{HyperFuture}~\cite{suris2021hyperbolic} & 2021 & 3D-ResNet~\cite{he2016deep} & MLP & \cite{shao2020finegym,marszalek2009actions} & RGB & -- & \xmark & feat. & \xmark & \checkmark\\ 

        AVT \cite{girdharAnticipativeVideoTransformer2021} & 2021 & TSN~\cite{wangTemporalSegmentNetworks2016},ViT~\cite{dosovitskiyImageWorth16x162021}\comm{,F. r-cnn~\cite{ren2015faster}} & TF & \cite{damen2018scaling,damen2022rescaling,li2018eye,stein2013combining} & RGB,O & score & \xmark & feat. & \xmark & \checkmark\\
        DCR~\cite{xuLearningAnticipateFuture2022}& 2022 & TSN~\cite{wangTemporalSegmentNetworks2016},TSM~\cite{lin2019tsm}\comm{,F. r-cnn~\cite{ren2015faster}} & LSTM,TF & \cite{damen2018scaling,damen2022rescaling,li2018eye,stein2013combining} & RGB,O,M & score & \xmark & feat. & \xmark & \xmark\\

        RAFTformer~\cite{girase2023latency} & 2023 & MViT~\cite{li2022mvitv2} & TF & \cite{damen2018scaling,damen2022rescaling,li2018eye}  & RGB & -- & \xmark & feat. & \xmark & \xmark \\

        \rev{ARR}~\cite{liu2024recognition} & 2024 & AIM~\cite{yang2023aim} & TF & \cite{damen2022rescaling,li2018eye,stein2013combining} & RGB & -- & \xmark & feat. & \xmark & \checkmark\\ 

        \midrule
        
        \multicolumn{11}{c}{\textbf{Utilizing long-term temporal history}} \\
        \midrule
        Gammulle\etal~\cite{gammulle2019forecasting}& 2019 & ResNet~\cite{he2016deep} & LSTM & \cite{stein2013combining,kuehneLanguageActionsRecovering2014} & RGB,A & mid & \xmark & feat. & \checkmark &\xmark\\

        TRN~\cite{xu2019trn} & 2019 & TS~\cite{xiong2016twostream}  & LSTM & \cite{thumos2014,geest2016online} & RGB,M & mid & \xmark & feat. & \xmark & \xmark \\
        
        TempAgg \cite{senerTemporalAggregateRepresentations2020}& 2020 & TSN~\cite{wangTemporalSegmentNetworks2016},I3D~\cite{carreiraQuoVadisAction2017}\comm{,F. r-cnn~\cite{ren2015faster}} & NLB,LSTM & \cite{damen2018scaling,stein2013combining,kuehneLanguageActionsRecovering2014} & RGB,O,BB,M & score & \xmark & feat. & \checkmark & \xmark\\
        TTPP~\cite{wangTTPPTemporalTransformer2020}& 2020 &VGG\cite{simonyan2014vgg},TS\cite{xiong2016twostream} & TF,MLP & \cite{geest2016online,patron2010high,thumos2014} & RGB,M & mid & \xmark & feat. & \xmark & \xmark\\

        LAP~\cite{qu2020lapnet} & 2020 & TS~\cite{xiong2016twostream} & GRU & \cite{thumos2014,geest2016online} & RGB,M & mid & \xmark & feat. & \xmark & \xmark \\
        
        ImagineRNN~\cite{wuLearningAnticipateEgocentric2021}& 2021 & TSN~\cite{wangTemporalSegmentNetworks2016}\comm{,F. r-cnn~\cite{ren2015faster}} & LSTM & \cite{damen2018scaling,li2018eye} & RGB,O,M & score & \xmark & feat. & \xmark & \xmark\\

        LSTR~\cite{xu2021lstr} & 2021 & TS~\cite{xiong2016twostream} & TF & \cite{thumos2014,geest2016online} & RGB,M & mid & \xmark & feat. & \xmark & \xmark \\
        
        OadTR~\cite{wang2021oadtr}& 2021 & TS~\cite{xiong2016twostream} & TF & \cite{geest2016online,thumos2014} & RGB,M & mid & \xmark & feat. & \xmark & \xmark\\
        HRO~\cite{liuHybridEgocentricActivity2022}& 2022 & TSN~\cite{wangTemporalSegmentNetworks2016}\comm{,F. r-cnn~\cite{ren2015faster}} & GRU & \cite{damen2018scaling,li2018eye} & RGB,O,M & score & \xmark & feat. & \xmark & \xmark\\
        MeMViT~\cite{wuMeMViTMemoryAugmentedMultiscale2022} & 2022 & MViT~\cite{li2022mvitv2} & -- & \cite{damen2022rescaling} & RGB & -- & \xmark & feat. & \xmark & \checkmark\\
        FUTR~\cite{gongFutureTransformerLongterm2022}& 2022 & I3D~\cite{carreiraQuoVadisAction2017} & TF & \cite{stein2013combining,kuehneLanguageActionsRecovering2014} & RGB,M & mid & \xmark & feat. & \checkmark & \xmark\\
        TeSTra~\cite{zhao2022realtime} & 2022 & TS~\cite{xiong2016twostream},TSN~\cite{wangTemporalSegmentNetworks2016} & TF & \cite{thumos2014,damen2022rescaling} & RGB,M & mid & \xmark & feat. & \xmark & \xmark \\

        \rev{IAM}~\cite{tai2022inductive} & 2022 & TSN~\cite{wangTemporalSegmentNetworks2016},\cite{liu2022convnet},Swin~\cite{liu2021swin} & TF,MLP & \cite{damen2018scaling,damen2022rescaling,li2018eye} & RGB & -- & \xmark & feat. & \xmark & \xmark\\

        \rev{MAT}~\cite{wang2023memory} & 2023 & TS~\cite{xiong2016twostream},TSN~\cite{wangTemporalSegmentNetworks2016} & TF & \cite{thumos2014,geest2016online,damen2022rescaling} & RGB,M & mid & \xmark & feat. & \xmark & \xmark\\ 
        
        \rev{JOADAA}~\cite{guermal2024joadaa} & 2024 & I3D~\cite{carreiraQuoVadisAction2017} & TF & \cite{thumos2014,yeung2018every,sigurdsson2016hollywood} & RGB,M & mid & \xmark & feat. & \xmark & \xmark  \\ 
        
        \rev{BiOMamba}~\cite{wang2025biomamba} & 2025 & TSN~\cite{wangTemporalSegmentNetworks2016} & Mamba & \cite{thumos2014,geest2016online} & RGB,M & mid & \xmark & feat. & \xmark & \xmark\\ 
        
        \rev{ScalAnt}~\cite{zhong2026scalable} & 2026 & TSN~\cite{wangTemporalSegmentNetworks2016} & Mamba,TF & \cite{damen2022rescaling,graumanEgo4DWorld0002022,thumos2014} & RGB & -- & \xmark & feat. & \checkmark & \xmark\\ 
        
        \midrule

        \multicolumn{11}{c}{\textbf{Multi-modal fusion}} \\
        \midrule
        Zhou\etal~\cite{zhou2015temporal}& 2015 & VGG~\cite{simonyan2014vgg},TS~\cite{simonyan2014twostream} & MLP & \cite{soomro2012ucf101,zhou2015temporal} & RGB,O,M & late & \xmark & feat. & \xmark & \xmark\\
        Mahmud\etal~\cite{mahmud2017joint}& 2017 & C3D~\cite{tran2015c3D} & LSTM,MLP & \cite{oh2011large,rohrbach2012database} & O,M & late & \xmark & feat. & \xmark & \xmark\\
        Shen\etal~\cite{shen2018egocentric} & 2018 & \cite{zhu2015handmask},SSD~\cite{liu2016ssd} & LSTM & \cite{fathi2011learning,fathi2012learning} & O,BB,G,H & late & \xmark & feat. & \xmark & \xmark\\
        
        Liang\etal~\cite{liangPeekingFuturePredicting2019} & 2019 & \cite{chen2018encoder}\comm{,Mask r-cnn~\cite{he2017mask}} & LSTM & \cite{awad2018trecvid} & RGB,HOI & late & \xmark & feat. & \xmark & \checkmark \\
        
        RULSTM~\cite{furnariWhatWouldYou2019}& 2019 & TSN~\cite{wangTemporalSegmentNetworks2016}\comm{,F. r-cnn~\cite{ren2015faster}} & LSTM & \cite{damen2018scaling,li2018eye} & RGB,O,M& score & \xmark & feat. & \xmark & \xmark\\
        FHOI~\cite{liuForecastingHumanObjectInteraction2020}& 2020 & I3D~\cite{carreiraQuoVadisAction2017},CSN~\cite{tran2019csn}\comm{,F. r-cnn~\cite{ren2015faster}} & FC & \cite{damen2018scaling,li2018eye} & RGB,O & score & \xmark & feat. & \xmark & \checkmark\\
        Ego-OMG~\cite{dessalene2021forecasting}& 2021 & CSN~\cite{tran2015c3D},\comm{I3D~\cite{carreiraQuoVadisAction2017},UNet~\cite{},}GCN~\cite{kipf2016gcn} & LSTM & \cite{damen2018scaling} & RGB & score & \xmark & feat. & \xmark & \xmark\\
        Zatsarynna\etal~\cite{zatsarynnaMultiModalTemporalConvolutional2021}& 2021 & TSN~\cite{wangTemporalSegmentNetworks2016}\comm{,F. r-cnn~\cite{ren2015faster}} & TCN & \cite{damen2018scaling,damen2022rescaling} & RGB,O,M & late & \xmark & feat. & \xmark & \xmark\\
        Roy\etal~\cite{royActionAnticipationUsing2021}& 2021 & \comm{Mask r-cnn~\cite{he2017mask},}I3D~\cite{carreiraQuoVadisAction2017} & TF & \cite{damen2018scaling,stein2013combining,kuehneLanguageActionsRecovering2014} & RGB,HOI,M & score & \xmark & feat. & \xmark & \xmark\\
        AFFT \cite{zhong2023afft} & 2023 & TSN~\cite{wangTemporalSegmentNetworks2016},Swin~\cite{liu2021swin}\comm{,F. r-cnn~\cite{ren2015faster}} & TF & \cite{damen2022rescaling,li2018eye} & RGB,O,M,Au & mid & \xmark & feat. & \xmark & \xmark\\
        
        \rev{InAViT}~\cite{roy2024interaction} & 2024 & MotionFormer~\cite{patrick2021keeping} & -- & \cite{damen2022rescaling,li2018eye} & RGB,HOI & mid & \xmark & feat. & \xmark & \checkmark\\ 
        
        \midrule

        \multicolumn{11}{c}{\textbf{Conditioning on extra information}} \\
        \midrule
        S-RNN~\cite{jain2016structural}& 2016 & \cite{koppula2013learning} & GNN,LSTM & \cite{koppula2013learning} & RGB,O,BB & mid & \xmark & feat. & \xmark & \xmark\\
        DR$^2$N~\cite{sunRelationalActionForecasting2019}& 2019 & S3D~\cite{xie2018s3d}\comm{,F. r-cnn~\cite{ren2015faster}} & GNN,GRU & \cite{gu2018ava} & RGB & -- & \xmark & feat.,act. & \xmark & \checkmark\\
        Farha\etal~\cite{abu2020long} & 2020 & I3D~\cite{carreiraQuoVadisAction2017} & TCN,GRU & \cite{stein2013combining,kuehneLanguageActionsRecovering2014} & RGB,M & mid & \xmark & feat. & \checkmark & \xmark \\
        Camporese\etal~\cite{camporese2020knowledge} & 2020 & TSN~\cite{wangTemporalSegmentNetworks2016}\comm{,F. r-cnn~\cite{ren2015faster}} & LSTM & \cite{damen2018scaling,li2018eye} & RGB,O,M & score & \checkmark & feat. & \xmark & \xmark \\
        
        RESTEP~\cite{li2021restep} & 2021 & I3D~\cite{carreiraQuoVadisAction2017}\comm{,F. r-cnn~\cite{ren2015faster}} & ConvGRU & \cite{gu2018ava} & RGB & -- & \xmark & feat. & \xmark & \xmark \\

        A-ACT~\cite{gupta2022act} & 2022 & TSN~\cite{wangTemporalSegmentNetworks2016},I3D~\cite{carreiraQuoVadisAction2017} & TF,MLP & \cite{kuehneLanguageActionsRecovering2014,stein2013combining,damen2018scaling}& RGB,O,M & score & \xmark & feat.,act. & \xmark & \xmark \\

        Abst. goal~\cite{roy2022predicting} & 2022 & TSN~\cite{wangTemporalSegmentNetworks2016}\comm{,F. r-cnn~\cite{ren2015faster}} & GRU & \cite{damen2018scaling,damen2022rescaling,li2018eye} & RGB,O,M & score & \xmark & feat. & \xmark & \xmark \\
        
        \textsc{Anticipatr} \cite{nawhal2022rethinking} & 2022 & I3D~\cite{carreiraQuoVadisAction2017} & TF & \cite{stein2013combining,kuehneLanguageActionsRecovering2014,li2018eye,damen2018scaling} & RGB & -- & \xmark & feat. & \checkmark & \xmark \\
        
        I-CVAE~\cite{mascaro2023intention} & 2023 & SlowFast~\cite{feichtenhofer2019slowfast} & MLP,VAE,TF & \cite{graumanEgo4DWorld0002022} & RGB & -- & \xmark & feat. & \checkmark & \xmark \\

        \rev{AntGPT}~\cite{zhao2024antgpt} & 2024 & CLIP~\cite{radford2021clip} & LLM & \cite{graumanEgo4DWorld0002022,damen2018scaling,li2018eye} & RGB & -- & \checkmark & act. & \checkmark & \xmark \\
        
        \rev{S-GEAR}~\cite{diko2024semantically} & 2024 & ViT~\cite{dosovitskiyImageWorth16x162021} & TF & \cite{damen2018scaling,damen2022rescaling,li2018eye,stein2013combining} & RGB,O,M & score & \xmark & feat. & \xmark & \checkmark\\ 
        
        \rev{PALM}~\cite{kim2024palm} & 2024 & EgoVLP~\cite{lin2022egocentric} & LLM & \cite{graumanEgo4DWorld0002022,damen2018scaling,li2018eye} & RGB & -- & \checkmark & act. & \checkmark & \xmark\\ 
        
        \rev{PlausiVL}~\cite{mittal2024can} & 2024 & ViT~\cite{dosovitskiyImageWorth16x162021} & LLM & \cite{graumanEgo4DWorld0002022,damen2022rescaling} & RGB & -- & \checkmark & act. & \checkmark & \xmark\\ 
        
        \rev{ActionLLM}~\cite{wang2025multimodal} & 2025 & I3D~\cite{carreiraQuoVadisAction2017} & LLM & \cite{stein2013combining,kuehneLanguageActionsRecovering2014} & RGB,A & mid & \checkmark & feat. & \checkmark & \xmark\\ 

        \rev{ICVL}~\cite{cao2025vision} & 2025 & CLIP~\cite{radford2021clip} & LLM & \cite{graumanEgo4DWorld0002022,damen2018scaling,li2018eye} & RGB & -- & \checkmark & feat.,act. & \checkmark & \checkmark\\ 
        
        \rev{INSIGHT}~\cite{chu2026intention} & 2026 & EgoVideo-V~\cite{pei2024egovideo} & LLM & \cite{graumanEgo4DWorld0002022,damen2018scaling,li2018eye} & RGB,HOI & mid & \checkmark & feat.,act. & \checkmark & \checkmark\\ 
        
        \rev{AGA}~\cite{tai2026action} & 2026 & TSN~\cite{wangTemporalSegmentNetworks2016},Swin~\cite{liu2021swin} & MLP & \cite{damen2022rescaling,damen2018scaling,li2018eye} & RGB & -- & \xmark & feat. & \xmark & \xmark\\ 

        \bottomrule
        
    \end{tabular}}
    \label{tab:summary_models}
\end{table*}

\begin{table*}[ht!]
    \centering
    \ContinuedFloat  
    \renewcommand*{\arraystretch}{1}
    \caption{\rev{Summary of most relevant action anticipation models (continued). Trajectory (Traj.)}}
    \setlength\tabcolsep{4pt}
    \resizebox{\linewidth}{!}{
    \begin{tabular}{@{}lcllllccccc@{}}
        \toprule
        \multirow{2}{*}{Method} & \multirow{2}{*}{Year} & \multirow{2}{*}{\shortstack[l]{Feature \\ extractor}} & \multirow{2}{*}{\shortstack[l]{Future\\ predictor}} & \multirow{2}{*}{Datasets} & \multicolumn{5}{c}{Details} \\ 
        \arrayrulecolor{gray} \cmidrule(lr){6-11} \arrayrulecolor{black}
        & & & & & Modalities & Fusion & EK & Abst. & LT &E2E \\ 
        \midrule
        
        \multicolumn{11}{c}{\textbf{Incorporating uncertainty}} \\
        \midrule
        Schydlo\etal~\cite{schydlo2018anticipation}& 2018 & \cite{sung2012unstructured} & LSTM & \cite{koppula2013learning} & P & -- & \xmark & feat. & \xmark & \xmark\\
        Farha\etal~\cite{abufarhaUncertaintyAwareAnticipationActivities2019} & 2019 & -- & GRU & \cite{stein2013combining,kuehneLanguageActionsRecovering2014} & A,T & mid & \xmark & act. & \checkmark & \xmark \\
        APP-VAE~\cite{mehrasaVariationalAutoEncoderModel2019}& 2019 & -- & LSTM,VAE & \cite{kuehneLanguageActionsRecovering2014,yeung2018every} & A,T & mid & \xmark & act. & \checkmark & \xmark\\
        Ng\etal~\cite{ng2020forecasting}& 2020 & I3D~\cite{carreiraQuoVadisAction2017} & GRU & \cite{sigurdsson2016hollywood,kuehneLanguageActionsRecovering2014,rohrbach2012database,stein2013combining} & RGB & -- & \xmark & feat. & \checkmark & \xmark\\
        AGG~\cite{piergiovanni2020adversarial} & 2020 & -- & TCN,MLP,GAN& \cite{sigurdsson2016hollywood,stein2013combining,yeung2018every} & A & -- & \xmark & act. & \checkmark & \xmark \\
        Zhao\etal~\cite{zhao2020diverse} & 2020 & -- & LSTM,GAN & \cite{kuehneLanguageActionsRecovering2014,stein2013combining,damen2018scaling} & A,T & mid & \xmark & act. & \checkmark & \xmark \\

        \rev{Guan\etal}~\cite{guan2020generative} & 2020 & TSN~\cite{wangTemporalSegmentNetworks2016}, ORB-SLAM~\cite{mur2015orb} & GRU,MLP & \cite{damen2018scaling} & RGB,Traj. & mid & \xmark & feat. & \xmark & \xmark\\ 
        
        \rev{DiffAnt}~\cite{zhong2023diffant} & 2023 & I3D~\cite{carreiraQuoVadisAction2017} & TF,Diff & \cite{kuehneLanguageActionsRecovering2014,stein2013combining,damen2018scaling,li2018eye} & RGB & -- & \xmark & feat. & \checkmark & \xmark\\ 
        
        \rev{UADT}~\cite{guo2024uncertainty} & 2024 & MViT~\cite{li2022mvitv2},TSN~\cite{wangTemporalSegmentNetworks2016} & TF & \cite{damen2022rescaling,li2018eye,stein2013combining} & RGB,O,M & mid & \xmark & feat. & \xmark & \xmark\\ 
        
        \rev{CPM}~\cite{xie2024towards} & 2024 & TSM~\cite{lin2019tsm} & TF & \cite{damen2022rescaling,damen2018scaling,li2018eye} & RGB,O,M & score & \xmark & feat. & \xmark & \xmark\\ 
        
        \rev{GTD}~\cite{zatsarynna2024gated} & 2024 & I3D~\cite{carreiraQuoVadisAction2017},TSM~\cite{lin2019tsm} & TCN,Diff & \cite{kuehneLanguageActionsRecovering2014,stein2013combining,sener2022assembly101} & RGB & -- & \xmark & act. & \checkmark & \xmark\\ 
        
        \rev{ActFusion}~\cite{gong2024actfusion} & 2024 & I3D~\cite{carreiraQuoVadisAction2017} & TF,Diff & \cite{stein2013combining,kuehneLanguageActionsRecovering2014,fathi2011learning} & RGB & -- & \xmark & act. & \checkmark & \xmark\\ 
        
        \rev{MANTA}~\cite{zatsarynna2025manta} & 2025 & I3D~\cite{carreiraQuoVadisAction2017},TSM~\cite{lin2019tsm} & Mamba,Diff & \cite{kuehneLanguageActionsRecovering2014,stein2013combining,sener2022assembly101} & RGB & -- & \xmark & act. & \checkmark & \xmark\\ 
        
        \rev{VEDiT}~\cite{lin2025vedit} & 2025 & SigLIP~\cite{zhai2023sigmoid} & DiT & \cite{graumanEgo4DWorld0002022} & RGB & -- & \xmark & feat. & \checkmark & \xmark\\ 
        
        \rev{MixANT}~\cite{wasim2025mixant} & 2025 & I3D~\cite{carreiraQuoVadisAction2017},TSM~\cite{lin2019tsm} & Mamba,Diff & \cite{kuehneLanguageActionsRecovering2014,stein2013combining,sener2022assembly101} & RGB & -- & \xmark & act. & \checkmark & \xmark\\ 
        \midrule

        \multicolumn{11}{c}{\textbf{Modeling at a more conceptual level}} \\
        \midrule
        Farha\etal~\cite{farhaWhenWillYou2018}& 2018 & -- & CNN,GRU & \cite{stein2013combining,kuehneLanguageActionsRecovering2014} & A,T & mid & \xmark & act. & \checkmark & \xmark\\
        Miech\etal~\cite{miechLeveragingPresentAnticipate2019} & 2019 & R(2+1)D~\cite{tran2018closer} & FC & \cite{damen2018scaling,kuehneLanguageActionsRecovering2014,caba2015activitynet} & RGB & -- & \xmark & feat.,act. & \xmark & \xmark\\
        Ke\etal~\cite{keTimeConditionedActionAnticipation2019}& 2019 & I3D~\cite{carreiraQuoVadisAction2017} & TCN & \cite{damen2018scaling,stein2013combining,kuehneLanguageActionsRecovering2014} & RGB,T & mid & \xmark & act. & \checkmark & \xmark\\

        Zhang\etal~\cite{zhang2020intuitionanalysis}& 2020 & TSN~\cite{wangTemporalSegmentNetworks2016} & LSTM,MLP & \cite{damen2018scaling} & RGB,O,M & score & \xmark & feat.,act. & \xmark & \xmark \\

        \textsc{ProActive}~\cite{gupta2022proactive} & 2022 & -- & TF & \cite{kuehneLanguageActionsRecovering2014,caba2015activitynet,yeung2018every} & A,T & mid & \checkmark & act. & \checkmark & \xmark \\


        \bottomrule
        
    \end{tabular}}
    \label{tab:summary_models}
\end{table*}

\subsection{Leveraging temporal dynamics as a learning signal}
\label{sec:learning_signal}

\textit{Utilizing temporal dynamics via a self-supervised loss.}
One promising way to equip machines with the ability to infer future user actions is to use the abundantly available unlabeled videos on the Internet. These videos are economically feasible to obtain at massive scales and contain rich signals, particularly the inherent temporal ordering of frames. Early models leveraged this temporal ordering by learning to predict future video frames~\cite{zeng2017visual} or motion images~\cite{rodriguez2018dynamic} directly in pixel space, and then used the predicted frames for downstream action classification. However, predicting in pixel space is inherently challenging due to its high dimensionality and extreme variability~\cite{opreaReviewDeepLearning2020}. 
\rev{
Notably, Zeng\etal~\cite{zeng2017visual} demonstrated that pixel-level synthesis was feasible on synthetic data such as Moving MNIST~\cite{srivastava2015unsupervised}, but it became computationally intractable and significantly less effective on natural videos, necessitating a shift to feature-space prediction. Although recent generative video models~\cite{openai2024sora,wan2025wan} have improved frame synthesis fidelity, generating high-fidelity frames as an intermediate step remains computationally impractical for anticipation and introduces additional error propagation through a subsequent classification stage. Consequently, the action anticipation community has converged on predicting future visual representations rather than raw pixels.
}

\rev{
This design choice is further corroborated by the broader self-supervised learning literature, where world models~\cite{bardes2024revisiting,assran2025v} have demonstrated that predicting in a learned feature space, rather than in pixel space, yields more robust and semantically meaningful representations~\cite{lecun2022path}.
} 
Following this paradigm, action anticipation methods learn to anticipate future representations by maximizing the correlation~\cite{fernandoAnticipatingHumanActions2021,teeti2023tdino} or minimizing the distance between predicted and actual future frame representations, e.g., in the form of a contrastive loss~\cite{oord2018infonce,han2019dpc,han2020medpc,suris2021hyperbolic,zatsarynna2022self,tan2023multiscale} or a simple MSE loss~\cite{vondrick2016anticipating,gaoREDReinforcedEncoderDecoder2017,zhong2018unsupervised,girdharAnticipativeVideoTransformer2021}.

Video representations capture the semantic information about actions and can be automatically computed using a pre-trained feature extractor, making them scalable to unlabeled videos. 
Vondrick\etal~\cite{vondrick2016anticipating} initiated this line by building a regression network that takes a single frame as input and anticipates the future representation of a pre-trained AlexNet~\cite{Alexnet2012}. To incorporate historical context, Zhong and Zheng~\cite{zhong2018unsupervised} developed a two-stream architecture: a spatial stream encodes the current frame, while a temporal stream summarizes the observation history with an LSTM~\cite{hochreiter1997long}. To mitigate the LSTM's limited long-range capacity, a fully connected layer aggregates all hidden states into a compact representation before predicting future features.

\textit{\rev{Contrastive learning.}}
Another line of research builds on contrastive learning~\cite{oord2018infonce,han2019dpc}. Contrastive Predictive Coding (CPC)~\cite{oord2018infonce} and Dense Predictive Coding (DPC)~\cite{han2019dpc} share a common predictive coding foundation, where the learning objectives are to predict coarse clip-level and fine-grained spatiotemporal region representations of future clips, respectively. Han\etal~\cite{han2020medpc} extended DPC by introducing learnable memory banks to account for the non-deterministic nature of predicting the future. Suris\etal~\cite{suris2021hyperbolic} further integrated hyperbolic geometry, specifically the Poincar\'{e} ball model, into DPC; the hyperbolic space encodes hierarchical structures~\cite{nickel2017poincare}, enabling confident predictions of specific actions and graceful fallback to higher-level abstractions under uncertainty.
Separately, Zatsarynna\etal~\cite{zatsarynna2022self} proposed a composite loss for unintentional action prediction, combining a temporal contrastive loss that encourages proximity of neighboring clips with a pair-wise ordering loss that captures the relative temporal order.

\textit{\rev{Combining self-supervised and supervised objectives.}}
In contrast to the aforementioned approaches, where the main model remains fixed after pre-training and only a classifier is trained, several works have demonstrated the benefit of fine-tuning or jointly training with self-supervised and supervised losses. In the two-stage pretrain-then-finetune paradigm, Gao\etal~\cite{gaoREDReinforcedEncoderDecoder2017} proposed an LSTM-based encoder-decoder (RED) that takes a sequence of historical visual representations and outputs anticipated future representations for classification. The first stage performs self-supervised training with an MSE loss on all training videos; the second stage fine-tunes with full supervision on positive segments. A reinforcement learning module was additionally introduced for sequence-level optimization (see Section~\ref{sec:extrainfo}). Xu\etal~\cite{xuLearningAnticipateFuture2022} adopted order-aware pre-training for a transformer-based reasoning model~\cite{vaswaniAttentionAllYou2017}, where the observed video is sent into the transformer without positional encoding and the temporal order is supervised via a Gaussian affinity over positional encodings~\cite{hayat2019gaussian}. Fine-tuning on the target dataset has proved particularly beneficial when positive segments constitute a small fraction of the data~\cite{teeti2023tdino,tan2023multiscale}.
Other works train with joint supervised and self-supervised objectives in a single stage. Tran\etal~\cite{tran2021knowledge} combined cross-entropy with an attentional-pooling-based feature distance loss to handle spatial dynamics in videos. Fernando and Herath~\cite{fernandoAnticipatingHumanActions2021} proposed bounded similarity measures that overcome the limitations of L2 distance and cosine similarity for correlating predicted and actual future representations. Girdhar and Grauman~\cite{girdharAnticipativeVideoTransformer2021} employed an MSE loss to supervise intermediate future predictions within a GPT-2-based causal model~\cite{radford2019language}. \rev{Building on a similar causal decoder architecture, Liu\etal~\cite{liu2024recognition} further introduced an unsupervised pre-training stage that leverages the natural temporal ordering of unlabeled video to predict next-frame features, providing a better initialization for the anticipation model.} In contrast to predicting the next frame feature using causal masking as in~\cite{girdharAnticipativeVideoTransformer2021}, Girase\etal~\cite{girase2023latency} proposed a generalized masking-based self-supervision scheme to predict shuffled future features, allowing for exponentially more variations than the vanilla causal masking scheme.
\rev{
Zhong\etal~\cite{zhong2023learning} replaced deterministic reconstruction with a diffusion process that models the distribution of a masked clip's representation conditioned on adjacent clip embeddings, jointly optimized with a video-text matching loss derived from CLIP~\cite{radford2021clip} pseudo labels.
}

\textit{Exploring the underlying dynamics.}
Some approaches relied on the adversarial training framework~\cite{goodfellow2014gan} to learn the underlying dynamics of videos~\cite{zeng2017visual,gammulle2019joint,piergiovanni2020adversarial,zhao2020diverse}. For instance, Zeng\etal~\cite{zeng2017visual} formulated the anticipation task as an inverse reinforcement learning (IRL) problem, where the goal was to imitate the behavior of natural sequences that are treated as expert demonstrations. More specifically, they leveraged the Generative Adversarial Imitation Learning framework~\cite{ho2016generative} to bypass the exhaustive state-action pair visits, since both the state (frame) and action (transition) space were high-dimensional and continuous. In this framework, a discriminator, which aimed to distinguish the generated sequence from expert’s, and a generator (policy), guided by the discriminator to move toward expert-like regions, were jointly optimized during training.
\subsection{Utilizing long-term temporal history}
\label{sec:history}
Most actions exhibit dependencies on preceding actions. These preceding \textit{trigger} actions may not necessarily be confined to the immediate past but can extend back to the long-term history. Consequently, many methods aim to exploit the information contained in the long-term history. 

\textit{Recurrent networks.} Having an internal memory, recurrent networks, including LSTMs~\cite{hochreiter1997long} and GRUs~\cite{chung2014gru}, are often adopted for exploiting the sequential action history~\cite{gaoREDReinforcedEncoderDecoder2017,furnariWhatWouldYou2019,xu2019trn,qu2020lapnet}.
A prominent design axis among RNN-based methods is the joint modeling of online detection and anticipation by feeding predicted future features back into the detection pipeline~\cite{xu2019trn,qu2020lapnet,wang2021oadtr}. The temporal recurrence network (TRN)~\cite{xu2019trn} pioneered this idea by simultaneously anticipating the immediate future and using the prediction to enhance present-moment detection. Qu\etal~\cite{qu2020lapnet} (LAP-Net) extended TRN by adaptively selecting the temporal sampling range via Gumbel-Softmax reparameterization~\cite{jang2016gumbelsoftmax}, conditioned on the inferred action progression. Other recurrent approaches tackle complementary challenges without relying on predicted future features. Qi\etal~\cite{qi2021selfregulated} addressed error accumulation in recursive prediction with a self-regulated framework using contrastive loss and dynamic frame reweighting.\comm{based on the cosine similarity.}
\rev{
Tai\etal~\cite{tai2022inductive} introduced a higher-order recurrent architecture (IAM) that queries an indexed memory of past hidden states via multi-head attention, gating the retrieved context with the current frame encoding.
}
\rev{
Beyond classical RNNs, BiOMamba~\cite{wang2025biomamba} adopted the Mamba state-space model~\cite{gu2024mamba} for joint detection and anticipation, compressing distant history into long-term memory and applying forward-backward temporal modeling so that later evidence can correct earlier representations earlier via backward replay.
}

\textit{Memory augmentation.} Recurrent networks struggle to model long-term dependencies due to their sequential nature, motivating a family of memory-augmented designs. At one end of the spectrum, \emph{learned external memories} store dataset-level knowledge: Gammulle\etal~\cite{gammulle2019forecasting} maintained an updatable neural memory that reads from and writes to observed frames and action labels, enriching the current-step representation; in contrast, Liu and Lam~\cite{liuHybridEgocentricActivity2022} (HRO) employed a fixed memory module to predict the next-step representation, complemented by contrastive regularization~\cite{wuLearningAnticipateEgocentric2021} and a past-to-future transition layer. At the other end, \emph{cached clip memories} enable end-to-end training over long histories without prohibitive cost. Wu\etal~\cite{wuMeMViTMemoryAugmentedMultiscale2022} (MeMViT) processed videos as sequential clips and cached intermediate key--value pairs from earlier steps, granting current queries access to distant frames without any learnable memory parameters.
\rev{
However, the cost of attention-based retrieval still grows with sequence length. Zhong\etal~\cite{zhong2026scalable} proposed the Cross Linear Attentive Memory (CLAM) module, which reformulates linear attention~\cite{katharopoulos2020transformerrnn,yang2024gla} as a cross-attention-style retrieval mechanism with linear computation and constant memory cost, selectively extracting frame-level context cues that are decoded by a non-autoregressive Transformer to predict multiple future actions in one shot.
}

\textit{Attention-based temporal modeling.} An alternative to sequential recurrence is to process the observed history in \textit{parallel} via attention~\cite{vaswaniAttentionAllYou2017}. Early approaches adopted full self- or cross-attention over the frame sequence: TTPP~\cite{wangTTPPTemporalTransformer2020} generated a query from the most recent frame to attend over all past key--value features, yielding an aggregated representation from which future actions were iteratively decoded; TempAgg~\cite{senerTemporalAggregateRepresentations2020} employed non-local blocks~\cite{wang2018nonlocal,wu2019featurebank} to aggregate short-term and long-term features at multiple temporal granularities, combining outputs via score fusion.
Because standard self-attention scales as $O(N^2)$ in the sequence length $N$, subsequent work focused on compressed or streaming formulations. Low-rank factorizations~\cite{choromanski2021performer,wang2020linformer} and query-based cross-attention~\cite{rae2020compressive,jaegle2021perceiver} offer general strategies to reduce complexity. For anticipation specifically, LSTR~\cite{xu2021lstr} introduced a two-stage memory compression that abstracts the long-term history into a fixed-length latent representation, upon which a short window of recent frames performs self- and cross-attention; learnable tokens appended to the short-term memory serve as anticipation queries.
\rev{
This encoder--decoder paradigm with explicit long-short-term memory separation inspired further works for joint detection and anticipation: MAT~\cite{wang2023memory} introduced iterative circular dependencies between memory and anticipated future representations, while Guermal\etal~\cite{guermal2024joadaa}, akin to TRN~\cite{xu2019trn} in the recurrent setting, fed the anticipated future embedding back to enrich the present-frame representation.
}
Zhao and Krähenbühl~\cite{zhao2022realtime} (TeSTra) pushed efficiency further by reformulating the first compression stage through the lens of temporal kernel smoothing~\cite{tsai2019TransformerDissection}, applying box and Laplace kernels to achieve streaming attention with constant per-frame caching and computation overhead.

\subsection{Narrowing anticipation space by using additional information}
\label{sec:narrow}

While vision based systems are the de-facto standard for action recognition and anticipation \cite{furnariWhatWouldYou2019, girdharAnticipativeVideoTransformer2021}, using other additional supporting input modalities like optical flow features~\cite{wangTemporalSegmentNetworks2016,carreiraQuoVadisAction2017} or knowledge about objects and sounds in the scene~\cite{furnari2017next,pasca2024summarize,kazakosEPICFusionAudioVisualTemporal2019} has shown to be beneficial~\cite{sun2022datamodalities}. To properly fuse the different modalities, various fusion strategies have been exploited for action anticipation in recent years, which are described in Section~\ref{sec:multimodal}. Moreover, additional useful information such as personalization (Section~\ref{sec:extrainfo}) could also be incorporated in the system to further support anticipation.

\subsubsection{Multi-modal fusion}
\label{sec:multimodal}

\textit{Input modalities.} While action anticipation methods typically take the original video frames as the default input modality, many methods leverage higher level modalities from the original RGB frames, such as optical flow~\cite{zach2007opticalflow}, object features (usually depending on an object detector such as \cite{ren2015faster,he2017mask}), human-object-interaction~\cite{qiLearningHumanObjectInteractions2018,royActionAnticipationUsing2021,roy2022interaction}, and human-scene-interaction~\cite{liangPeekingFuturePredicting2019}. For egocentric videos, the camera wearer's trajectory~\cite{park2016egocentric}, hand trajectory~\cite{liuForecastingHumanObjectInteraction2020}, eye gaze~\cite{li2018eye}, and environment affordance~\cite{nagarajan2020ego,mur2024aff} have been additionally used.

\begin{figure*}[t]
   \centering
     \begin{subfigure}{.3\linewidth}
         \centering
         \includegraphics[width=\linewidth]{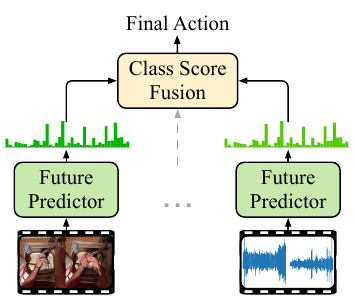}
         \caption{Score fusion.}
         \label{fig:score-fusion}
     \end{subfigure}\hfill
     \begin{subfigure}{.265\linewidth}
         \centering
         \vspace{0.2cm}
         \includegraphics[width=\linewidth]{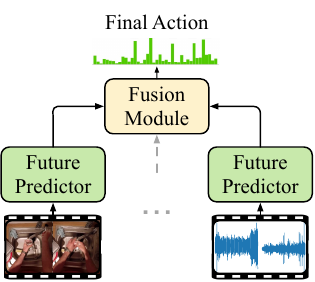}
         \caption{Late feature fusion.}
         \label{fig:late-fusion}
     \end{subfigure}\hfill
     \begin{subfigure}{.255\linewidth}
         \centering
         \vspace{0.2cm}
         \includegraphics[width=\linewidth]{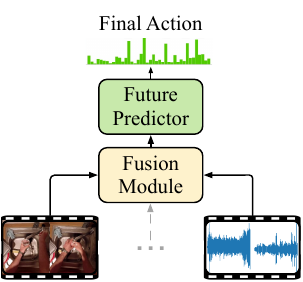}
         \caption{Mid-level feature fusion.}
         \label{fig:mid-level-fusion}
     \end{subfigure}
     \caption{Three typical fusion strategies for multi-modal action anticipation. Feature extractors are omitted for brevity. Examples from the EpicKitchens~\cite{damen2018scaling} dataset and AFFT~\cite{zhong2023afft}.} \vspace{-2mm}
     \label{fig:fusion}
\end{figure*}

These additional modalities can be grouped into several categories.
\textit{(i)~Gaze and hand cues:} Shen\etal~\cite{shen2018egocentric} used gaze transitions to extract events and modeled their inter-relationships with a Hawkes process~\cite{hawkes1971spectra}; Liu\etal~\cite{liuForecastingHumanObjectInteraction2020} (FHOI) jointly predicted hand trajectories, interaction hotspots, and future action labels; and Dessalene\etal~\cite{dessalene2021forecasting} used hand-object contact time as an anticipatory cue, combined with a GCN~\cite{kipf2016gcn} for long-term temporal modeling.
\textit{(ii)~Human-object interaction (HOI):} Roy and Fernando~\cite{royActionAnticipationUsing2021} represented HOI via pairwise cross-correlation of visual features and aggregated them with an encoder-decoder transformer.
\rev{Roy\etal~\cite{roy2024interaction} proposed InAViT, which refines interaction-centric tokens through spatial and trajectory cross-attention and fuses them with video tokens for a richer anticipatory representation.}
\textit{(iii)~Language as auxiliary modality:}
\rev{Pasca\etal~\cite{pasca2024summarize} proposed TransFusion, which encodes compact textual summaries of past actions with SBERT~\cite{reimers2019sentence} and fuses them with multi-scale visual features via joint attention.}
Detailed per-method modality configurations are given in Table~\ref{tab:summary_models}.

\textit{Score fusion.} Consistent with most multi-modal action recognition models \cite{wangTemporalSegmentNetworks2016,carreiraQuoVadisAction2017}, anticipation models typically use score fusion (Fig.~\ref{fig:score-fusion}) to combine modalities. Fixed-weight averaging~\cite{senerTemporalAggregateRepresentations2020,royActionAnticipationUsing2021,girdharAnticipativeVideoTransformer2021} already improves over uni-modal baselines, but Furnari\etal~\cite{furnariWhatWouldYou2019} demonstrated that dynamic modality weighting is particularly beneficial: their RU-LSTM architecture learns attention scores over appearance, motion, and object-presence branches via fully connected fusion layers. Osman\etal~\cite{osman2021slowfast} extended this framework to multi-rate streams inspired by~\cite{feichtenhofer2019slowfast}.

\textit{Feature fusion.} An alternative line of work fuses modality representations before the final decision, following either a late or mid-level strategy (Fig.~\ref{fig:late-fusion},~\ref{fig:mid-level-fusion}).
\textit{Late fusion} methods first predict modality-specific future representations independently and then combine them: Mahmud\etal~\cite{mahmud2017joint} jointly learned action labels and starting times from three branches fused via a fully-connected layer, while Zatsarynna\etal~\cite{zatsarynnaMultiModalTemporalConvolutional2021} predicted future representations with per-modality TCNs~\cite{lea2017tcn} before concatenation and fusion.
\textit{Mid-level fusion} methods merge modalities before temporal anticipation. Standard approaches concatenate and project multi-modal features~\cite{gongFutureTransformerLongterm2022,abufarhaUncertaintyAwareAnticipationActivities2019} \rev{or use simple summation~\cite{mur2024aff}}, whereas Zhong\etal~\cite{zhong2023afft} (AFFT) replaced hand-designed fusion with a transformer module using a modality-agnostic token akin to the \texttt{[cls]} token in ViT~\cite{dosovitskiyImageWorth16x162021}, outperforming score-fusion counterparts. Notably, AFFT also incorporated audio but found it less informative than visual modalities. The GPT-based anticipation module from~\cite{girdharAnticipativeVideoTransformer2021} was used for final prediction.

\subsubsection{Conditioning on extra information}
\label{sec:extrainfo}

\textit{Personalization.} Zhou and Berg~\cite{zhou2015temporal} explored two simple tasks related to temporal prediction in egocentric videos of everyday activities, pairwise temporal ordering and future video selection. They showed that personalization to a particular individual or environment provides significantly increased performance.

\textit{Prediction time.} As the recurrent approaches suffer from increasing inference time and error accumulation when predicting longer action sequences, some methods relied on parallel decoding by learning a sequence of action queries~\cite{wang2021oadtr,gongFutureTransformerLongterm2022,nawhal2022rethinking}. However, the number of predictable future actions was also limited to the number of action queries used in the training process. Ke\etal~\cite{keTimeConditionedActionAnticipation2019} took another approach and chose to condition on a time variable representing the prediction time. Specifically, they transformed the prediction time to a time representation, and concatenated it with the original inputs forming time-conditioned observations. Their model was therefore capable of anticipating a future action at arbitrary and variable time horizons in a one-shot fashion. Building upon a similar concept, \textsc{Anticipatr}~\cite{nawhal2022rethinking} utilized a linear layer to convert learnable action queries, along with the anticipation duration, into time-conditioned queries. 
It employed a two-stage learning approach. In the first stage, a segment encoder is trained to predict the set of future action labels. In the second stage, a video encoder and an anticipation decoder were trained for making the final decision based on the input of the segment encoder. 

\textit{Sequence-level information.} Standard cross-entropy losses operate per time step and do not capture sequence-level quality. To address this, Gao\etal~\cite{gaoREDReinforcedEncoderDecoder2017} introduced a reinforcement learning reward that favors early correct predictions, while Ng and Fernando~\cite{ng2020forecasting} reweighted the classification loss to down-weight short observations and emphasize near-future correctness. Orthogonally, cycle-consistency modules~\cite{abu2020long,gupta2022act} predict past activities from the projected future as an auxiliary constraint, and Fosco\etal~\cite{fosco2023etm} proposed the Event Transition Matrix (ETM), an explicit prior over action-to-action transition likelihoods computed from dataset statistics.

\textit{Semantic meaning of actions.} Most anticipation methods rely on one-hot labels, which may not capture the underlying distributional structure of future actions. Camporese\etal~\cite{camporese2020knowledge} addressed this by injecting semantic similarity via word-embedding-based (GloVe~\cite{pennington2014glove}) label smoothing. \rev{
More recently, Diko\etal~\cite{diko2024semantically} proposed S-GEAR, which transfers geometric associations among action prototypes from language space to visual space through temporal context aggregation and prototype alignment blocks. Tai\etal~\cite{tai2026action} took a complementary approach with Action-Guided Attention (AGA), constructing queries and keys from predicted action distributions to retrieve relevant past moments based on semantic action dependencies rather than pixel-level features.
}

\rev{
\textit{Language and vision-language models.}
The recent success of large language models (LLMs)~\cite{gpt5.4,gemini3,opus4.6} and vision-language models (VLMs)~\cite{bai2025qwen25vl,bai2025qwen3} has opened a new avenue for action anticipation by providing rich commonsense and procedural knowledge.
Existing approaches can be grouped into three architectural subfamilies.
\textit{(i)~Text-interface LLM methods} represent video observations as discrete action sequences and prompt LLMs: AntGPT~\cite{zhao2024antgpt} combines bottom-up and top-down (goal-inference) prompting with Llama2~\cite{touvron2023llama2}/GPT-3.5~\cite{chatgpt} and further distills the LLM into a compact student at 1.3\% of the original size; PALM~\cite{kim2024palm} enriches the textual context with VLM-generated captions and selects diverse in-context exemplars via maximal marginal relevance.
\textit{(ii)~Vision-language fusion methods} tighten the coupling between visual and textual modalities: ActionLLM~\cite{wang2025multimodal} uses Llama~\cite{touvron2023llama} as the backbone and fuses action cues through a Cross-Modality Interaction Block; ICVL~\cite{cao2025vision} mitigates the information loss of text-only approaches by fusing VLM-inferred intention embeddings~\cite{grattafiori2024llama3} with visual features via cross-attention and fine-tuning Llama3 with LoRA~\cite{hu2022lora}, complemented by multi-modal exemplar selection.
\textit{(iii)~Plausibility-aware modeling:} PlausiVL~\cite{mittal2024can} trains a video-language model to distinguish plausible from counterfactual futures through a plausibility loss, with a complementary repetition loss to penalize degenerate action sequences.
}

\textit{Spatio-temporal graph structure.} Human-object interactions unfold in both space and time, motivating spatio-temporal graph models whose nodes represent scene entities and edges encode their interactions~\cite{jain2016structural,qiLearningHumanObjectInteractions2018,sunRelationalActionForecasting2019}. The key design choice lies in how node features are updated. Jain\etal~\cite{jain2016structural} used per-node LSTMs over concatenated messages; Qi\etal~\cite{qiLearningHumanObjectInteractions2018} and Sun\etal~\cite{sunRelationalActionForecasting2019} learned adaptive adjacency weights via attention, akin to Graph Attention Networks~\cite{velivckovic2017gat}, isolating spatial and temporal dependencies by pairing the graph with a GRU~\cite{chung2014gru}. Li\etal~\cite{li2021restep} argued that such isolation ignores spatio-temporal co-occurrence and proposed a ConvGRU-based~\cite{ballas2015convgru} relational network (STRN) that jointly captures both.

\begin{figure}[t]
    \centering
    \includegraphics[width=0.6\linewidth]{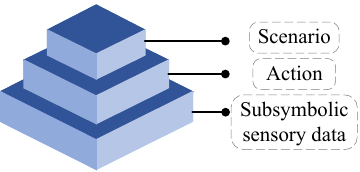} \vspace{-1mm}
    \caption{Abstraction level of activities.}\vspace{-2mm}
    \label{fig:abstraction_level}
\end{figure}

\textit{The underlying intention.} To get something done, humans perform a sequence of actions dictated by an intention~\cite{wu2017anticipating,kruglanski2020habitual}. It is therefore advantageous to understand human intention to anticipate how a person would act in the future. Following this idea, some works~\cite{roy2022latentgoal,roy2022predicting,mascaro2023intention,zhao2024antgpt} built a \textit{top-down} framework (refer to Fig.~\ref{fig:abstraction_level}) to explicitly infer the intention and use the intention to constrain the variability of future actions. Specifically, Roy and Fernando~\cite{roy2022latentgoal} used a stacked LSTM to derive the latent goal from the observed visual representations, and exploited the concepts of \textit{goal closeness} and \textit{goal consistency} to guide the predictions. Goal closeness suggests actions should align with the latent goal, while goal consistency maintains this alignment over a sequence.
Mascaro\etal~\cite{mascaro2023intention} proposed estimating human intention with an MLP Mixer~\cite{tolstikhin2021mlp}. By taking the past actions, the estimated intention, as well as a latent vector sampled from a latent distribution based on the reparameterization trick from VAE as input, a transformer decoder was used to generate a sequence of future actions. 
Zatsarynna and Gall~\cite{zatsarynna2023goal} integrated goal estimation into a multi-task framework, demonstrating improved performance compared to the counterpart that lacks goal estimation. 
\rev{
Chu\etal~\cite{chu2026intention} proposed INSIGHT, a two-stage framework that combines semantically enriched perception with explicit cognitive reasoning. The first stage extracts hand-object interaction representations and corrects action predictions using a verb-noun co-occurrence matrix, while the second stage formulates anticipation as a structured \textit{think} $\rightarrow$ \textit{reason} $\rightarrow$ \textit{answer} process, where a reinforcement-learning-based module explicitly infers user intention before forecasting future actions with Qwen2.5-VL~\cite{bai2025qwen25vl}.
}
\subsection{Incorporating uncertainty}
\label{sec:uncertainty}

Predicting future actions is inherently uncertain. Given an observed video segment containing an ongoing action, multiple actions could be possible to be the next action following the observed one. This uncertainty becomes even larger if we are going to predict far into the future. Therefore, it may be beneficial to model the underlying uncertainty, allowing to capture different possible future actions. However, action prediction is mostly treated as a classification problem and optimized using the cross-entropy loss, suffering from overly high resemblance to dominant ground truth while suppressing other reasonable possibilities~\cite{dai2017towards}. Moreover, approaches that are optimized using the mean square error tend to produce the mean of the modes~\cite{vondrick2016anticipating,mathieu2016deep}. Therefore, some approaches output multiple possible future actions to be able to capture the underlying uncertainty.

\textit{Generating multiple outputs with multiple rules.} To address the non-deterministic nature of future prediction, Vondrick\etal~\cite{vondrick2016anticipating} extended their regression network to support multiple outputs by training a mixture of $K$ networks, where each mixture was trained to predict one of the modes in the future. As only one of the possible futures is provided in the dataset and it is unknown to which of the $K$ mixtures each data sample belongs, they overcame these issues by alternating between two steps: 1) optimizing the network while keeping the mixture assignment fixed, and 2) re-estimating the mixture assignment with the new network weights. During inference, the most likely action class is selected by marginalizing over the probability distributions of all modes. Following a similar idea, Piergiovanni\etal~\cite{piergiovanni2020adversarial} proposed a differentiable grammar model which learns a set of production rules, being able to generate multiple candidate future sequences that follow a similar distribution of sequences seen during training. To avoid enumerating all possible rules which have exponential growth in time, they introduced adversarial learning for the grammar model, allowing for much more memory- and computationally-efficient learning without such enumeration.

\textit{Generating multiple outputs by sampling from the learned distribution.} A family of methods generates diverse predictions by sampling from learned distributions at inference time. Schydlo\etal~\cite{schydlo2018anticipation} sampled from an encoder-decoder LSTM's action probability distribution at each step and applied beam search to select final candidates. Farha and Gall~\cite{abufarhaUncertaintyAwareAnticipationActivities2019} extended this idea to joint prediction of action labels and durations using a two-model architecture~\cite{farhaWhenWillYou2018}, modeling the former with cross-entropy and the latter with a Gaussian likelihood, and generating long-term sequences by recursive autoregressive sampling. Zhao and Wildes~\cite{zhao2020diverse} further improved joint label-time anticipation by treating both as discrete categorical variables~\cite{li2017time} and training a conditional adversarial \textit{seq2seq} generator~\cite{sutskever2014seq2seq} with Gumbel-Softmax relaxation~\cite{jang2016gumbelsoftmax} for differentiable sampling and a diversity regularizer~\cite{yang2019diversity}. Instead of adversarial training, Mehrasa\etal~\cite{mehrasaVariationalAutoEncoderModel2019} employed a recurrent VAE~\cite{kingma2013auto} with a time-varying prior that captures temporal dependencies between successive actions, overcoming the limitation of a fixed $\mathcal{N}(0,I)$ prior.
\rev{Guan\etal~\cite{guan2020generative} proposed an invertible generative model for joint trajectory-action forecasting from egocentric video, also using Gumbel-Softmax~\cite{jang2016gumbelsoftmax} for discrete actions while enabling exact likelihood computation to balance diversity and precision. 
Abdelsalam\etal~\cite{abdelsalam2023gepsan} adopted a conditional VAE on top of a transformer context encoder to sample multiple plausible next procedural steps in natural language.
}

\rev{
\textit{Probabilistic consistency and uncertainty-aware prediction.} A complementary line of work uses probabilistic reasoning not to diversify outputs but to improve the internal consistency of the prediction process. Xie\etal~\cite{xie2024towards} proposed a Consistency-guided Probabilistic Model (CPM) that enforces temporal continuity by defining past-aware and global-aware distributions over the full sequence and optimizing the evidence lower bound to reduce their divergence. Guo\etal~\cite{guo2024uncertainty} proposed an uncertainty-aware decoupling framework (UADT) that splits anticipation into verb-to-noun and noun-to-verb sub-problems and uses each stream's predictive uncertainty to select cross-stream features and dynamically fuse predictions.
}

\rev{
\textit{Generating multiple outputs via iterative denoising.} More recently, diffusion models~\cite{ho2020ddpm} have emerged as a powerful generative paradigm for stochastic action anticipation. Existing methods differ in the space where denoising is performed. In the \emph{embedding space}, DiffAnt~\cite{zhong2023diffant} iteratively denoises Gaussian noise into future action embeddings conditioned on observed video, while VEDiT~\cite{lin2025vedit} applies a diffusion Transformer~\cite{peebles2023scalable} with a flow-matching noise schedule~\cite{esser2024scaling} in the embedding space of frozen pretrained visual encoders. In the \emph{action probability space}, Gated Temporal Diffusion (GTD)~\cite{zatsarynna2024gated} models observed and unobserved parts as a unified sequence with gated temporal convolutions steering information flow. ActFusion~\cite{gong2024actfusion} extends this paradigm by introducing anticipative masking that replaces future frames with learnable mask tokens, unifying temporal action segmentation and anticipation within a single joint denoising process and revealing bidirectional benefits between the two tasks. Subsequent works advance the temporal backbone: MANTA~\cite{zatsarynna2025manta} replaces gated convolutions with bidirectional Mamba-style~\cite{gu2024mamba} state-space layers for global receptive fields at linear cost, and MixANT~\cite{wasim2025mixant} further introduces mixture-of-experts state-transition matrices for adaptive temporal memory propagation.
}
\subsection{Modeling at a more conceptual level}
\label{sec:concept}

Most action anticipation methods introduced in previous sections operate on the feature-level, i.e., the anticipation module receives a single or a sequence of feature vectors representing the observed video, and then predicts the future feature vectors that are further categorized to actions by a classifier. Some approaches first recognize the ongoing action(s) and then infer future actions directly from the observed action history. By operating on discrete action symbols rather than dense feature vectors, these methods expose interpretable action-transition patterns and are especially natural for long-term anticipation, where symbolic action sequences can offer a complementary representation to dense visual features.

\textit{Markov assumption.} To this end, some methods make a Markov assumption on the sequence of performed actions and use a linear layer to model the transition between the past and the next action~\cite{miechLeveragingPresentAnticipate2019,zhang2020intuitionanalysis,fernandoAnticipatingHumanActions2021}.
The linear weights of the transition model can be interpreted as conveying the importance of each past action class for the anticipation. By analyzing the linear weights and the confidence vector of the current action, we can easily interpret which action class is most responsible for the anticipation and diagnose the source of false predictions, i.e., whether the false prediction is due to the recognition model or due to the learned transitional model. 

\textit{Long-term anticipation.} The action-level event history provides a complementary perspective to feature vectors for modeling how activities evolve, enabling predictions farther into the future~\cite{gammulle2019forecasting,mehrasaVariationalAutoEncoderModel2019}. Consequently, many approaches utilize such action-level history to perform long-term anticipation~\cite{farhaWhenWillYou2018,zhao2020diverse,mehrasaVariationalAutoEncoderModel2019}. For instance, Farha\etal~\cite{farhaWhenWillYou2018} introduced two methods: an RNN model that recursively predicted the next action class and its duration, and a CNN model that produced the entire future action sequence as a matrix in one step.

\textit{Continuous-time event modeling.} Continuous-time action sequences can also be viewed as sparse events and modeled with temporal point processes (TPPs)~\cite{kingman1992poisson,hawkes1971spectra}. An initial effort by Mahmud\etal~\cite{mahmud2016poisson} adopted a Poisson process with a Gaussian Process prior over the intensity function for activity inter-arrival time modeling. Mehrasa\etal~\cite{mehrasaVariationalAutoEncoderModel2019} extended this with a learned, observation-dependent intensity and VAE-based stochastic sampling. Gupta and Bedathur~\cite{gupta2022proactive} replaced intensity estimation altogether with temporal normalizing flows~\cite{rezende2015variational,mehrasa2019pointprocessflow} conditioned on sequence dynamics, enabling closed-form sampling and additionally addressing goal prediction and goal-conditioned sequence generation.

\rev{
\textit{Connection to LLM/VLM-based methods.} Among the LLM- and VLM-based methods discussed in Section~\ref{sec:extrainfo}, the text-interface approaches~\cite{zhao2024antgpt,kim2024palm} most closely align with this conceptual-level paradigm, as they reason entirely over discrete action labels; others~\cite{mittal2024can,cao2025vision,chu2026intention} additionally incorporate continuous visual features. In contrast, the preceding approaches in this section require no external pretrained language model and instead operate directly on recognized action symbols and their transition structure.
}

\subsection{\rev{Relationship to anticipation horizons}}
\rev{
The method categories described in earlier sections are organized by technical contribution rather than by anticipation horizon, but clear tendencies emerge (see the ``LT'' column in Table~\ref{tab:summary_models}). Methods based on self-supervised temporal dynamics (Section~\ref{sec:learning_signal}) and multi-modal fusion (Section~\ref{sec:multimodal}) are used predominantly for short-term anticipation, where dense feature-level prediction and immediate cues such as motion or objects are most informative. By contrast, long-term anticipation more often relies on higher-level modeling, including action-level reasoning (Section~\ref{sec:concept}), uncertainty-aware prediction (Section~\ref{sec:uncertainty}), and knowledge- or language-based conditioning (Section~\ref{sec:extrainfo}), since distant futures are both more abstract and more ambiguous. Methods that exploit extended temporal history (Section~\ref{sec:history}) span both settings: they improve short-term prediction by enlarging the observation window, but can also be adapted to decode sequences of future actions for long-term anticipation. Overall, short-term anticipation is largely perception-driven, whereas long-term anticipation places greater emphasis on abstraction, uncertainty, and semantic priors.
}
 \section{Evaluation Datasets and Metrics}

\subsection{Datasets}
The algorithms for action anticipation are evaluated on a wide range of datasets that are temporally labeled with the corresponding action. To better differentiate current datasets, we consider several characteristics in Table~\ref{tab:datasets}: spontaneity of behavior (scripted vs.\ natural), viewpoint (egocentric, third-person, or multi-view), presence of composite activities (elementary actions forming higher-level scenarios, cf.\ Fig.~\ref{fig:abstraction_level}), and concurrent activities (simultaneous performance of multiple actions). Below, we describe the most widely used benchmarks.


\newcolumntype{P}[1]{>{\centering\arraybackslash}p{#1}}
\newcolumntype{M}[1]{>{\centering\arraybackslash}m{#1}}

\begin{table*}[t]
    \centering
    \caption{Representative benchmark datasets for human action anticipation. The following abbreviations are used in the table. Type: Activities of daily living (ADL). Sensor modality: Depth (D), Au (Audio). Comp. act.: Composite activities. Conc. act.: Concurrent activities. Spont.: Spontaneity.}
    \setlength\tabcolsep{5.5pt}
    \begin{tabular}{@{}lcccM{0.7cm}M{0.7cm}M{0.7cm}ccccl@{}}
        \toprule
        Dataset & Year & Type & View & Multi-view & Comp. act. & Conc. act. & Spont. & Hours & Segments & Classes & Sensor modality\\
        \midrule
        TV-I~\cite{patron2010high} & 2010 & Movie & Shooting & \xmark & \xmark & \xmark & Medium & -- & 300 & 4 & RGB\\
        50Salads~\cite{stein2013combining} & 2013 & Cooking & Top-view & \xmark & \checkmark & \xmark & Medium & 4.5 & 966 & 17 & RGB \\
        CAD-120~\cite{koppula2013learning} & 2013 & ADL & Shooting& \xmark & \checkmark & \xmark & Low & 0.57 & 120 & 12 & RGB, D \\
        Breakfast~\cite{kuehneLanguageActionsRecovering2014} & 2014 & Cooking & Shooting & \checkmark & \checkmark & \xmark & Medium & 77 & 11.2K & 48 & RGB \\
        THUMOS14~\cite{thumos2014} & 2014 & Web & Shooting & \xmark & \xmark & \xmark & Medium & 20 & 6.3K & 20 & RGB \\
        Charades~\cite{sigurdsson2016hollywood} & 2016 & ADL & Shooting & \xmark & \xmark & \checkmark & Low & 82.3 & 66.5K & 157 & RGB \\
        TVSeries~\cite{geest2016online} & 2016 & Movie & Shooting & \xmark & \xmark & \xmark & Medium & 16 & 6.2K & 30 & RGB \\
        EGTEA Gaze+\cite{li2018eye} & 2018 & Cooking & Egocentric & \xmark & \xmark & \xmark & Medium & 28 & 10.3K & 106 & RGB, Gaze \\
        EpicKitchens-50\cite{damen2018scaling} & 2018 & Cooking & Egocentric & \xmark & \xmark & Few & High & 55 & 39.6K & 2,513 & RGB, Au \\
        AVA~\cite{gu2018ava} & 2018 & Movie & Shooting & \xmark & \xmark & \checkmark & Medium & 107.5 & 38.6K & 60 & RGB \\
        EpicKitchens-100~\cite{damen2022rescaling} & 2020 & Cooking & Egocentric & \xmark & \xmark & \checkmark & High & 100 & 90K & 3,807 & RGB, Au \\
        Ego4D~\cite{graumanEgo4DWorld0002022} & 2022 & ADL & Egocentric & \xmark & \xmark & \checkmark & High & 243 & -- & 4,756 & RGB, Au\\
        Assembly101~\cite{sener2022assembly101} & 2022 & ADL & Multi-view & \checkmark & \checkmark & \xmark & Medium & 513 & 1M & 1,380 & RGB, Pose\\
        
        \rev{MECCANO}~\cite{ragusa2023meccano} & 2023 & ADL & Egocentric & \xmark & \xmark & \checkmark & Medium & 7.0 & 8.9K & 61 & RGB, D, Gaze \\
        \bottomrule
    \end{tabular}
    \label{tab:datasets}
\end{table*}

\textit{Web \& Movie datasets.}
Several widely used benchmark datasets are collected from the web or movies~\cite{patron2010high,thumos2014,geest2016online,gu2018ava}. 
\textbf{THUMOS14}~\cite{thumos2014} is a popular benchmark for temporal action detection and anticipation~\cite{gaoREDReinforcedEncoderDecoder2017,wangTTPPTemporalTransformer2020,wang2021oadtr,zhao2022realtime}. It contains over 20 hours of sport videos annotated with 20 actions. Since the training set contains only trimmed videos that cannot be used to train action anticipation models, prior works usually use the validation set (including 3K action instances in 200 untrimmed videos) for training and the test set (including 3.3K action instances in 213 untrimmed videos) for evaluation.
\textbf{TVSeries}~\cite{geest2016online} contains video footage from six popular TV series, about 150 minutes for each and about 16 hours in total. The dataset totally includes 30 realistic, everyday actions (e.g., \textit{pick up}, \textit{open door}, etc.), and every action occurs at least 50 times in the dataset.

\textit{Cooking activities.} 
One popular category of datasets in this domain contains videos of cooking activities: Epic-Kitchens~\cite{damen2018scaling,damen2020epic}, EGTEA~Gaze+~\cite{li2018eye}, Breakfast~\cite{kuehneLanguageActionsRecovering2014}, and 50Salads~\cite{stein2013combining}. 
Among these, Epic-Kitchens and EGTEA~Gaze+ are egocentric (first-person) datasets. The videos are recorded with a wearable camera that captures the scene directly in front of the user, in which only hands are visible in the center of the camera view. \textbf{EpicKitchens-100} consists of 700 long unscripted videos of cooking activities totalling 100 hours. It contains 90K action annotations, 97 verbs, and 300 nouns. Considering all unique (\textit{verb}, \textit{noun}) pairs in the public training set yields 3,806 unique actions. \textbf{EpicKitchens-55} is an earlier version of the EpicKitchens-100 with 39,596 segments labeled with 125 verbs, 352 nouns, and 2,513 combinations (actions), totalling 55 hours. \textbf{EGTEA Gaze+} contains 28 hours of videos including 10.3K action annotations, 19 verbs, 51 nouns, and 106 unique actions.
Breakfast~\cite{kuehneLanguageActionsRecovering2014} and 50Salads~\cite{stein2013combining} contain cooking activities in which subjects follow a specific recipe, resulting in actions performed without hesitation or mistakes (reduced spontaneity). 
The \textbf{Breakfast} dataset comprises 1,712 videos of 52 different individuals making breakfast in 18 different kitchens, totalling 77 hours. Every video is categorized into one of the 10 activities related to breakfast preparation. 
The videos are annotated by 48 fine-grained actions.  
The \textbf{50Salads} dataset comprises 50 top-view videos of 25 people preparing a salad. The dataset contains over 4 hours of RGB-D video data, annotated with 17 fine-grained action labels and 3 high-level activities.

\textit{Other daily living activities.} 
\textbf{Ego4D}~\cite{graumanEgo4DWorld0002022} is the most extensive daily-life egocentric video dataset. The forecasting benchmark from Ego4D consists of 120 hours of annotated videos from 53 different scenarios. The annotations provided contain 478 noun types and 115 verb types, with a total amount of 4756 action classes among training and validation set. The number of training and validation samples has doubled in the second version.
\textbf{Assembly101}~\cite{sener2022assembly101} features 4321 videos of participants assembling and disassembling 101 toy vehicles without fixed guidelines, resulting in varied action sequences, mistakes, and corrections. Captured from 8 static (RGB) and 4 egocentric (monochrome) cameras, it comprises 513 hours of footage with over 100K coarse and 1M fine-grained segments across 202 coarse and 1380 fine-grained classes. \rev{\textbf{MECCANO}~\cite{ragusa2023meccano} offers a complementary perspective: 20 participants assemble a single toy motorbike from 49 components, recorded in 20 egocentric videos with aligned RGB, depth, and gaze signals. It provides annotations for 12 verbs, 20 object classes, and 61 action classes across 8,857 temporal segments.}

\subsection{Evluation metrics}
\label{sec:metric}
\textbf{Top-k accuracy} is widely used to evaluate the overall performance of action anticipation methods. It is computed by checking if the ground truth label is among the top-k predictions. Accounting for the class imbalance in a long-tail distribution, many methods tend to use class-aware measures such as \textbf{Class Mean Top-k Accuracy} to evaluate the performance. Note that some works~\cite{furnariLeveragingUncertaintyRethink2018,furnariWhatWouldYou2019} refer to this metric as \textbf{Mean Top-k Recall}. 
In the case of long-term anticipation, where predictions for a large number of frames are made, some works~\cite{farhaWhenWillYou2018,keTimeConditionedActionAnticipation2019,senerTemporalAggregateRepresentations2020,gongFutureTransformerLongterm2022} report the class mean top-1 accuracy of the predicted frames (\textbf{Mean over Classes (MoC)}) as the evaluation metric.

In the case of multiple step predictions, \textbf{Mean Average Precision (mAP)}~\cite{yeung2018every} and its augmented version \textbf{Calibrated Average Precision (cAP)}~\cite{geest2016online} are widely used. mAP averages the per-class average precision across all action classes. cAP extends mAP by re-weighting precision with the ratio of negative to positive frames, thereby correcting for class imbalance so that precision is computed as if positive and negative frames were equally represented.

As treating predictions for each future time-step independently when calculating accuracy does not account for the sequential nature of the prediction task where the order of predictions is important, \textbf{Edit Distance (ED)}, computed as the Damerau-Levenshtein distance~\cite{damerau1964ed,levenshtein1966ed}, is employed to evaluate the predicted action sequences~\cite{graumanEgo4DWorld0002022}. The goal of this measure is to assess performance in a way which is robust to some error in the predicted order of future actions. A predicted verb/noun is considered "correct" if it matches the ground truth verb label at a specific timestep. The allowed operations to compute the edit distance are insertions, deletions, substitutions and transpositions of any two predicted actions. The lower the ED is the more similar are the anticipated sequences to the ground-truth.

\section{Benchmark Protocols and Results}

\subsection{Short-term anticipation}
\begin{table}[t]
\centering
\setlength\tabcolsep{3pt}
\caption{Comparison of the state-of-the-art methods addressing predictions at multiple timestamps on TVSeries~\cite{geest2016online} (mean cAP \%) and THUMOS-14~\cite{thumos2014} (mAP \%). The optimal performance in each column within every block is indicated in bold, while the supreme overall performance in each column is further denoted by an underscore.}
\label{tab:tvseries_thumos14}
 \resizebox{\linewidth}{!}{
    \begin{tabular}{@{}lcccccccccc@{}}
    \toprule
    \multirow{2}{*}{Method} & \multirow{2}{*}{Year} & \multirow{2}{*}{\shortstack{Back- \\ bone}} & \multirow{2}{*}{Pre-train} & \multirow{2}{*}{\rev{\makecell{Obs. \\ len.}}} & \multicolumn{3}{c}{TVSeries} & \multicolumn{3}{c}{THUMOS14} \\ \cmidrule(lr){6-8} \cmidrule(lr){9-11}
     & & & & & 1.0 & 2.0 & Avg. & 1.0 & 2.0 & Avg. \\
    \midrule
    ED~\cite{gaoREDReinforcedEncoderDecoder2017} & 2017 & \multirow{3}{*}{VGG} & \multirow{3}{*}{IN1K} & 4s & 68.8 & 66.7 & 68.7 & -- & -- & -- \\
    RED~\cite{gaoREDReinforcedEncoderDecoder2017} & 2017 & & & 4s & 70.2 & 66.8 & 69.4 & -- & -- & --\\
    TTPP\cite{wangTTPPTemporalTransformer2020} & 2020 & & & 2s & \textbf{71.6} & \textbf{69.3} & \textbf{71.3} & -- & -- & -- \\
    \midrule
    
    ED~\cite{gaoREDReinforcedEncoderDecoder2017} & 2017 & \multirow{11}{*}{TS} & \multirow{11}{*}{ANet1.3} & 4s & 74.6 & 71.0 & 74.5 & 36.8 & 31.6 & 36.6\\
    RED~\cite{gaoREDReinforcedEncoderDecoder2017} & 2017 & & & 4s & 75.5 & 71.2 & 75.1 & 37.5 & 32.1 & 37.5\\
    TRN~\cite{xu2019trn} & 2019 & & & 16s & 75.9 & 72.3 & 75.7 & 39.1 & 34.3 & 38.9\\
    TTPP\cite{wangTTPPTemporalTransformer2020} & 2020 & & & 2s & 77.6 & 74.9 & 77.9 & 41.0 & 37.3 & 40.9\\
    LAP~\cite{qu2020lapnet}  & 2020 & & & 16s & \textbf{78.9} & \textbf{75.5} & 78.7 & 43.2 & 37.0 & 42.6\\
    OadTR\cite{wang2021oadtr} & 2021 & & & 63s & 78.2 & 74.3 & 77.8 & 46.8 & 41.1 & 45.9\\
    LSTR~\cite{xu2021lstr}  & 2021 & & & 520s & -- & -- & 80.8 & -- & -- & 50.1\\
    TeSTra~\cite{zhao2022realtime} & 2022 & & & 520s & -- & -- & -- & 55.7 & 47.8 & 55.3 \\
    \rev{MAT}~\cite{wang2023memory} & 2023 & & & 264s & -- & -- & 81.5 & -- & -- & 57.3 \\
    \rev{MiniROAD}~\cite{an2023miniroad} & 2023 & & & -- & -- & -- & -- & 57.4 & 47.7 & 56.3 \\
    \rev{BiOMamba}~\cite{wang2025biomamba} & 2025 & & & 264s & -- & -- & \textbf{82.9} & \textbf{58.9} & \textbf{50.6} & \textbf{58.5} \\
    \midrule
    TTPP~\cite{wangTTPPTemporalTransformer2020}& 2020 & \multirow{9}{*}{TS} & \multirow{9}{*}{K400} & 2s & -- & -- & -- & 43.6 & 38.7 & 42.8 \\
    LSTR~\cite{xu2021lstr} & 2021 & & & 520s & -- & -- & -- & 53.3 & 45.7 & 52.6 \\
    OadTR\cite{wang2021oadtr} & 2021 & & & 63s & \textbf{\ul{80.1}} & \textbf{\ul{75.7}} & 79.1 & 54.6 & 46.8 & 53.5\\
    TeSTra~\cite{zhao2022realtime} & 2022 & & & 520s & -- & -- & -- & 57.4 & 48.9 & 56.8 \\
    \rev{MAT}~\cite{wang2023memory} & 2023 & & & 264s & -- & -- & 82.6 & -- & -- & 58.2 \\
    \rev{MiniROAD}~\cite{an2023miniroad} & 2023 & & & -- & -- & -- & -- & 60.7 & 51.4 & 59.7 \\
    \rev{JOADAA}~\cite{guermal2024joadaa} & 2024 & & & 136s & -- & -- & -- & \textbf{\ul{62.9}} & -- & -- \\
    \rev{BiOMamba}~\cite{wang2025biomamba} & 2025 & & & 264s & -- & -- & \textbf{\ul{83.7}} & 60.2 & 51.5 & 59.7 \\
    \rev{ScalAnt}~\cite{zhong2026scalable} & 2026 & & & -- & -- & -- & -- & 62.5 & \textbf{\ul{52.5}} & \textbf{\ul{61.6}} \\
    \bottomrule
    \end{tabular}} \vspace{-2mm}
\end{table}
Short-term action anticipation can be broadly classified into two types: future action predictions at multiple timestamps and single future action prediction. The former frequently employs datasets that incorporate third-person perspectives, such as TVSeries~\cite{geest2016online} and THUMOS14~\cite{thumos2014} (Section~\ref{sec:tvseries_thumos14}), while the latter is typically evaluated using egocentric datasets such as EpicKitchens~\cite{damen2022rescaling} (Section~\ref{sec:ek100}). \rev{An important distinction between these two evaluation settings lies in the task definition: TVSeries and THUMOS14 allow the prediction of ongoing actions (i.e., actions that have already begun at the time of prediction), whereas EpicKitchens requires the prediction of actions that have not yet started, making it a strictly anticipatory task.}

\subsubsection{TVSeries and THUMOS14}
\label{sec:tvseries_thumos14}
Table~\ref{tab:tvseries_thumos14} showcases the performance of approaches that address predictions at multiple timestamps, with results detailed at times 1.0 and 2.0 seconds. Comprehensive results spanning 0.25 to 2.0 seconds can be found in the supplementary material. The chosen evaluation metrics for TVSeries and THUMOS14 are mean calibrated average precision (mcAP) and mean average precision (mAP), respectively, as detailed in Section~\ref{sec:metric}. The backbones range from ImageNet-1K (IN1K) pre-trained models like VGG-16~\cite{simonyan2014vgg} to two-stream video models~\cite{xiong2016twostream} (TS), pre-trained either on ActivityNet~\cite{caba2015activitynet} (ANet1.3) or Kinetics~\cite{carreiraQuoVadisAction2017} (K400). 

Two key trends emerge from the results. First, the scale and domain of pre-training data have a substantial impact on performance. Upgrading from VGG-16 with ImageNet-1K to two-stream models pre-trained on ActivityNet consistently boosts results on TVSeries (e.g., TTPP: $71.3\% \rightarrow 77.9\%$). \rev{Pre-training on the larger Kinetics dataset yields further gains, as illustrated by BiOMamba, which improves from $82.9\%$ to $83.7\%$ average mcAP on TVSeries. A similar pattern can also be observed on THUMOS14, where Kinetics pre-training lifts ScalAnt to the best average mAP of $61.6\%$.
Second, architectural advances have driven considerable progress. Within each pre-training block, more recent methods that employ advanced temporal modeling, whether transformer-based (MAT, JOADAA, etc.), SSM-based (BiOMamba), or hybrid RNN-transformer designs (ScalAnt), consistently outperform earlier approaches. On THUMOS14 with ActivityNet pre-training, BiOMamba achieves $58.5\%$ average mAP, significantly outperforming earlier methods like TRN ($38.9\%$). With Kinetics pre-training, JOADAA obtains the highest single-timestamp mAP of $62.9\%$ at 1.0 seconds on THUMOS14. The diversity of architectures among the top-performing methods suggests that the key driver is the capacity for effective temporal modeling rather than any single architectural paradigm.
}

\subsubsection{EpicKitchens-100}
\label{sec:ek100}
\begin{table}[t]
    \centering
    \setlength\tabcolsep{1pt}
    \caption{Comparison of state-of-the-art methods on the validation and test set of EpicKitchens-100~\cite{damen2022rescaling} in terms of mean top-5 recall (\%). The optimal performance in each column within every block is indicated in bold. Additional modalities to the RGB modality: Objects (O), Bounding Boxes (BB), Motion (M), Audio (Au). \rev{Observation length is reported either in seconds (s) or in frames (f).}}
    \resizebox{\linewidth}{!}{
    \begin{tabular}{@{}clcccccccc>{\columncolor{Gray}}c@{}}
        \toprule
         \multirow{2}{*}{} & \multirow{2}{*}{Method} & \multirow{2}{*}{Year} & \multirow{2}{*}{\shortstack{Addl. \\ modality}} & \multirow{2}{*}{Backbone} & \multirow{2}{*}{Init.} & \multirow{2}{*}{E2E} & \multirow{2}{*}{\rev{\makecell{Obs. \\ len.}}} & \multicolumn{3}{c}{Overall}\\ 
          \cmidrule(lr){9-11}
          & & & & & & & & Verb & Noun & Act. \\ \midrule
         
         \multirow{27}{*}{\rotatebox{90}{Val}} 
          & RULSTM~\cite{furnariWhatWouldYou2019} & 2019 & -- & TSN & IN1K & \xmark & 2.75s & 27.5 & 29.0 & 13.3 \\
          & TempAgg \cite{senerTemporalAggregateRepresentations2020} & 2020 & -- & TSN & IN1K & \xmark & 6s & 24.2 & 29.8 & 13.0 \\
          & AVT~\cite{girdharAnticipativeVideoTransformer2021} & 2021 & -- & TSN & IN1K & \xmark & 10s & 27.2 & 30.7 & 13.6 \\
          & AVT~\cite{girdharAnticipativeVideoTransformer2021} & 2021 & -- & ViT-B & IN21K & \checkmark & 10s & 30.2 & 31.7 & 14.9 \\
          & MeMViT \cite{wuMeMViTMemoryAugmentedMultiscale2022} & 2022 & -- & MViT-B & K400 & \checkmark & 70.4s & 32.8 & 33.2 & 15.1 \\
          & MeMViT \cite{wuMeMViTMemoryAugmentedMultiscale2022} & 2022 & -- & MViT-L & K700 & \checkmark & 70.4s & 32.2 & 37.0 & 17.7 \\
          & DCR~\cite{xuLearningAnticipateFuture2022} & 2022 & -- & TSM & K400 & \xmark & 10s & 32.6 & 32.7 & 16.1 \\
          & RAFT~\cite{girase2023latency} & 2023 & -- & MViT-B & K400 & \xmark & 16s & 33.3 & 35.5 & 17.6 \\
          & RAFT~\cite{girase2023latency} & 2023 & -- & MViT-B & K700 & \xmark & 16s & 33.7 & 37.1 & 18.0 \\
          & \rev{S-GEAR}~\cite{diko2024semantically} & 2024 & -- & ViT-B & IN21K & \checkmark & 15s & 31.1 & 37.3 & 18.3 \\
          & \rev{ARR}~\cite{liu2024recognition} & 2024 & -- & AIM & -- & \checkmark & 8s & 31.3 & 32.0 & 16.3 \\
          & \rev{CPM}~\cite{xie2024towards} & 2024 & -- & TSM & K400 & \xmark & 30f & -- & -- & 17.2 \\
          & \rev{UADT}~\cite{guo2024uncertainty} & 2024 & -- & MViT-L & K700 & \xmark & -- & 38.2 & 41.4 & 20.3 \\
          & \rev{PlausiVL}~\cite{mittal2024can} & 2024 & -- & ViT-G/14 & -- & \checkmark & 32f & \textbf{55.6} & \textbf{54.2} & \textbf{27.6} \\
          & \rev{ScalAnt}~\cite{zhong2026scalable} & 2026 & -- & TSN & IN1K & \xmark & 32s & 31.1 & 27.2 & 17.6 \\
          & \rev{AGA}~\cite{tai2026action} & 2026 & -- & TSN & IN1K & \xmark & 30s & 32.2 & 35.7 & 17.5 \\
          & \rev{AGA}~\cite{tai2026action} & 2026 & -- & Swin-B & K400 & \xmark & 30s & 32.5 & 38.7 & 18.8 \\

         \arrayrulecolor{lightgray}\hhline{*{1}{~}*{10}{-}}\arrayrulecolor{black} \rule{0pt}{2.6ex}
         
          & RULSTM \cite{furnariWhatWouldYou2019} & 2019 & O,M & TSN & IN1K & \xmark & 2.75s & 27.8 & 30.8 & 14.0 \\
          & TempAgg~\cite{senerTemporalAggregateRepresentations2020} & 2020 & O,M,BB & TSN & IN1K & \xmark & 6 & 23.2 & 31.4 & 14.7 \\
          & AVT~\cite{girdharAnticipativeVideoTransformer2021} & 2021 & O & ViT-B & IN21K & \checkmark & 10s & 28.2 & 32.0 & 15.9 \\
          & DCR~\cite{xuLearningAnticipateFuture2022} & 2022 & O & TSM & K400 & \xmark & 10s & -- & -- & 18.3 \\
          & AFFT~\cite{zhong2023afft} & 2023 & O,M & TSN & IN1K & \xmark & 18s & 21.3 & 32.7 & 16.4 \\
          & AFFT~\cite{zhong2023afft} & 2023 & O,M,Au & Swin-B & K400 & \xmark & 16s & 22.8 & 34.6 & 18.5 \\
          & \rev{InAViT}~\cite{roy2024interaction} & 2024 & HOI & MotionFormer & -- & \checkmark & 2s & \textbf{52.5} & \textbf{51.9} & \textbf{25.9} \\
          & \rev{S-GEAR}~\cite{diko2024semantically} & 2024 & O & ViT-B & IN21K & \checkmark & 15s & 29.5 & 37.8 & 18.9 \\
          & \rev{UADT}~\cite{guo2024uncertainty} & 2024 & O,M & MViT-L & K700 & \xmark & -- & 43.5 & 46.6 & 23.0 \\
          & \rev{CPM}~\cite{xie2024towards} & 2024 & O,M & TSM & K400 & \xmark & 30f & -- & -- & 19.4 \\

         \midrule
         
         \multirow{11}{*}{\rotatebox{90}{Test}} 
          & RULSTM \cite{furnariWhatWouldYou2019} & 2019 & O,M & TSN & IN1K & \xmark & 2.75s & 25.3 & 26.7 & 11.2 \\
          & TempAgg \cite{senerTemporalAggregateRepresentations2020} & 2020 & O,M,BB & TSN & IN1K & \xmark & 6s & 21.8 & 30.6 & 12.6 \\
          & AVT \cite{girdharAnticipativeVideoTransformer2021} & 2021 & O & ViT-B & IN21K & \checkmark & 10s & 25.6 & 28.8 & 12.6 \\
          & TCN-TBN \cite{zatsarynnaMultiModalTemporalConvolutional2021} & 2021 & O,M & TBN & IN1K & \xmark & 5.25s & 21.5 & 26.8 & 11.0 \\
          & Abst. goal~\cite{roy2022predicting} & 2022 & O,M & TSN & IN1K & \xmark & 2s & 31.4 & 30.1 & 14.3 \\
          & AFFT~\cite{zhong2023afft} & 2023 & O,M,Au & Swin-B & K400 & \xmark & 16s & 20.7 & 31.8 & 14.9 \\
          & RAFT~\cite{girase2023latency} & 2023 & -- & MViT-B & K400 & \xmark & 16s & 27.3 & 32.8 & 14.0 \\
          & RAFT~\cite{girase2023latency} & 2023 & -- & MViT-B & K700 & \xmark & 16s & 27.4 & 34.0 & 14.7 \\
          & \rev{InAViT}~\cite{roy2024interaction} & 2024 & HOI & MotionFormer & -- & \checkmark & 2s & \textbf{49.1} & \textbf{50.0} & \textbf{23.8} \\
          & \rev{S-GEAR}~\cite{diko2024semantically} & 2024 & O & ViT-B & IN21K & \checkmark & 15s & 25.9 & 32.0 & 14.7 \\
          & \rev{AGA}~\cite{tai2026action} & 2026 & -- & Swin-B & K400 & \xmark & 30s & 30.8 & 36.4 & 16.9 \\

         \bottomrule
    \end{tabular}} \vspace{-2mm}
    \label{tab:ek100_sota_short}
\end{table}

Table~\ref{tab:ek100_sota_short} provides a detailed study of various state-of-the-art methods on the validation and test sets of EpicKitchens-100~\cite{damen2022rescaling}. It shows the \textit{overall} results, while results for \textit{unseen kitchen} and \textit{tail classes} are available in the supplementary material. Each of these benchmarks is evaluated based on the top-5 recall for verbs, nouns, and actions across all classes. The primary metric used to rank the methods is accentuated. If the anticipation approach employs a backbone fine-tuned for feature extraction, it is marked with a check mark in the column titled \textit{E2E} in Table~\ref{tab:ek100_sota_short}. 

\rev{
On the validation set, three observations can be made. First, advanced temporal modeling remains important. Among methods built on the same TSN backbone, AGA~\cite{tai2026action} (17.5) and ScalAnt~\cite{zhong2026scalable} (17.6) consistently outperform earlier approaches like AVT~\cite{girdharAnticipativeVideoTransformer2021} (13.6). 
Second, model scale and end-to-end training are clearly beneficial. PlausiVL~\cite{mittal2024can} achieves the strongest overall result with a single RGB stream (27.6), likely benefiting from its large-scale ViT-G/14 backbone, large language model, and end-to-end training, while other strong performers such as UADT~\cite{guo2024uncertainty}, AGA~\cite{tai2026action}, S-GEAR~\cite{diko2024semantically}, and RAFTformer~\cite{girase2023latency} further suggest the advantage of larger Transformer-based backbones and stronger pre-training over conventional architectures. 
Third, richer input modalities remain highly effective. By incorporating additional object and motion modalities, UADT~\cite{guo2024uncertainty} improves action recall from 20.3 to 23.0.
}

In the context of the test set, \rev{short challenge technical reports without full methodological descriptions are excluded, as they often employ ensembles of multiple methods or combine the training and validation splits, making fair comparison difficult.}
\rev{
InAViT~\cite{roy2024interaction} achieves the highest score ($23.8$), surpassing the previous best, AFFT~\cite{zhong2023afft} with the Swin-B backbone ($14.9$), by a large margin. Among RGB-only methods, AGA~\cite{tai2026action} with Swin-B attains the best performance ($16.9$), outperforming RAFTformer~\cite{girase2023latency} ($14.7$). Validation-set rankings largely hold on the test set, with InAViT and AGA maintaining their relative positions.
}

\subsection{Long-term anticipation}

\rev{
Long-term anticipation approaches are primarily evaluated on three benchmarks, Breakfast, 50Salads, and Ego4D, which differ in scale, viewpoint, and evaluation protocol. Breakfast and 50Salads are third-person datasets of procedural activities evaluated with mean-over-classes accuracy, whereas Ego4D is a large-scale egocentric benchmark evaluated with edit distance. Additionally, Breakfast and 50Salads require prediction of action duration, while Ego4D only requires prediction of a sequence of actions.
}
\subsubsection{Breakfast and 50Salads}
\label{sec:breakfast_50salads}
\begin{table}[t]
    \centering
    \caption{Benchmark of long-term action anticipation on Breakfast~\cite{kuehneLanguageActionsRecovering2014} and 50Salads~\cite{stein2013combining} in mean over classes ($\alpha = 0.3$). For details on the font coding, please refer to Table~\ref{tab:tvseries_thumos14}.}
    \label{tab:breakfast_50salads}
    \resizebox{.85\linewidth}{!}{
    \begin{tabular}{clccccc}
        \toprule
        \multirow{2}{*}{\makecell{Input \\ Type}} & \multirow{2}{*}{Method} & \multirow{2}{*}{Year} & \multicolumn{2}{c}{Breakfast $\beta$} & \multicolumn{2}{c}{50Salads $\beta$} \\
        \cmidrule(lr){4-5} \cmidrule(lr){6-7}
        & & & 0.2 & 0.5 & 0.2 & 0.5 \\
        \midrule
        \multirow{6}{*}{\rotatebox{90}{GT. label}}
        & RNN~\cite{farhaWhenWillYou2018} & 2018 & 50.25 & 41.75 & 29.51 & 10.38 \\
        & CNN~\cite{farhaWhenWillYou2018} & 2018 & 50.14 & 40.51 & 24.78 & 14.05 \\
        & Time-Cond.~\cite{keTimeConditionedActionAnticipation2019} & 2019 & 55.94 & 44.23 & 34.80 & 13.84 \\
        & UAAA~\cite{abufarhaUncertaintyAwareAnticipationActivities2019} (avg.) & 2019 & 42.94 & 33.07 & 24.65 & 14.34 \\
        & TempAgg~\cite{senerTemporalAggregateRepresentations2020} & 2020 & 56.10 & 41.50 & 32.70 & 15.30 \\
        & Zhao\etal~\cite{zhao2020diverse} (avg.) & 2020 & \textbf{\ul{71.32}} & \textbf{\ul{52.38}} & \textbf{\ul{36.37}} & \textbf{19.45} \\
        
        \midrule
        
        \multirow{11}{*}{\rotatebox{90}{Features}}
        & CNN~\cite{farhaWhenWillYou2018} & 2018 & 16.87 & 14.09 & - & - \\
        & TempAgg~\cite{senerTemporalAggregateRepresentations2020} & 2020 & 26.30 & 21.20 & - & - \\
        & Cycle Cons.~\cite{abu2020long} & 2020 & 27.37 & 25.20 & 23.70 & 15.89 \\
        & A-ACT~\cite{gupta2022act} & 2022 & 28.30 & 25.80 & 25.30 & 16.30 \\
        & FUTR~\cite{gongFutureTransformerLongterm2022} & 2022 & 29.88 & 25.87 & 24.86 & 15.26 \\
        & \rev{DiffAnt}~\cite{zhong2023diffant} & 2023 & 31.83 & \textbf{30.77} & \textbf{30.14} & 20.23 \\
        & \rev{ActFusion}~\cite{gong2024actfusion} & 2024 & 31.76 & 28.78 & 27.11 & \textbf{\ul{22.07}} \\
        & \rev{GTD}~\cite{zatsarynna2024gated} (avg.) & 2024 & 26.80 & 24.20 & 18.50 & 10.60 \\
        & \rev{ActionLLM}~\cite{wang2025multimodal} & 2025 & 29.45 & 24.55 & 29.06 & 18.37 \\
        & \rev{MANTA}~\cite{zatsarynna2025manta} (avg.) & 2025 & 30.90 & 27.70 & 21.90 & 13.00 \\
        & \rev{MixANT}~\cite{wasim2025mixant} (avg.) & 2025 & \textbf{32.80} & 28.70 & 23.70 & 14.60 \\
        \bottomrule
    \end{tabular}} \vspace{-2mm}
\end{table}

Table~\ref{tab:breakfast_50salads} reports long-term anticipation results on the Breakfast~\cite{kuehneLanguageActionsRecovering2014} and 50Salads~\cite{stein2013combining} datasets using mean over classes accuracy (see Section~\ref{sec:metric}), averaged over future timestamps within a defined anticipation duration. Actions are predicted after observing the first part ($\alpha$) of a video, with benchmarks from\cite{farhaWhenWillYou2018} setting $\alpha$ at 0.2 or 0.3. Predictions then span segments $\beta$ of the entire video, with $\beta = \{ 0.1, 0.2, 0.3, 0.5\}$. \rev{Methods that generate multiple outputs (Section~\ref{sec:uncertainty}) report the mean (avg.) over all outputs or the best-matching output (Top-1).} Due to space limits, only results with $\alpha = 0.3$ \rev{and avg.\ in case of multiple outputs} are shown in Table~\ref{tab:breakfast_50salads}. Complete results are in the supplementary material.

\rev{
Findings from the Breakfast and 50Salads datasets reveal several notable trends.
First, action recognition quality remains the primary bottleneck for long-term anticipation on Breakfast. Methods receiving ground truth action labels as input vastly outperform those operating on learned features: the best ground truth-based method (Zhao\etal~\cite{zhao2020diverse}: $71.32\%$ at $\beta{=}0.2$) more than doubles the best feature-based result (MixANT~\cite{wasim2025mixant}: $32.80\%$). This gap suggests that advances in upstream recognition could translate directly into anticipation improvements.
Second, on 50Salads, learned features have begun to close, and occasionally surpass, the ground truth gap. DiffAnt~\cite{zhong2023diffant} approaches the best ground truth-based result at $\beta{=}0.2$ ($30.14\%$ vs.\ $36.37\%$), while ActFusion~\cite{gong2024actfusion} exceeds it at $\beta{=}0.5$ ($22.07\%$ vs.\ $19.45\%$). Although the ground truth-label baselines date from 2020, partially confounding the comparison with algorithmic progress, these results indicate that learned spatiotemporal features can carry complementary information, such as inter-class dependencies and sub-action cues, that one-hot labels discard~\cite{abu2020long}.
}
The pronounced regularity of 50Salads action sequences, which facilitates coherent action-ordering learning~\cite{farhaWhenWillYou2018}, likely amplifies this effect.
\rev{
Third, generative models have substantially advanced feature-based anticipation. Recent diffusion-based approaches~\cite{zhong2023diffant, gong2024actfusion, zatsarynna2024gated, zatsarynna2025manta, wasim2025mixant} effectively exploit the richer semantic content of features and establish new state-of-the-art results on both datasets. LLM-based methods such as ActionLLM~\cite{wang2025multimodal} also show competitive accuracy, particularly on 50Salads.
}

\subsubsection{Ego4D}
\begin{table}[t]
    \centering
    \caption{\rev{Benchmark of long-term action anticipation on Ego4D~\cite{graumanEgo4DWorld0002022} in edit distance (lower is better). The best results are marked in bold.}}
    \label{tab:sota_ego4d}
    \setlength\tabcolsep{3pt}
    \resizebox{\linewidth}{!}{
    \begin{tabular}{lllccccc}
    \toprule
    Split & Method & Year & Backbone & LLM & Act. $\downarrow$ & Verb $\downarrow$ & Noun $\downarrow$ \\
    \midrule
    \multirow{8}{*}{\rotatebox{90}{Val-v1}}
    & SlowFast~\cite{feichtenhofer2019slowfast} & 2019 & SlowFast & -- & -- & 0.745 & 0.779 \\
    & RepLAI~\cite{mittal2022learning} & 2022 & R(2+1)D & -- & -- & 0.755 & 0.834 \\
    & I-CVAE~\cite{mascaro2023intention} & 2023 & SlowFast & -- & -- & 0.741 & 0.739 \\
    & HierVL~\cite{ashutosh2023hiervl} & 2023 & FrozenInTime & -- & -- & 0.723 & 0.734 \\
    & VideoLLM~\cite{chen2023videollm} & 2023 & -- & GPT2 & -- & 0.721 & 0.725 \\
    & VideoLLaMA~\cite{zhang2023video} & 2023 & ViT-G & Llama2-7B & -- & 0.703 & 0.721 \\
    & AntGPT~\cite{zhao2024antgpt} & 2024 & EgoVLP & Llama2-7B & -- & 0.700 & 0.717 \\
    & PlausiVL~\cite{mittal2024can} & 2024 & ViT-G & Llama2-7B & -- & \textbf{0.679} & \textbf{0.681} \\
    
    \midrule
    \multirow{3}{*}{\rotatebox{90}{Val-v2}}
    & PALM~\cite{kim2024palm} & 2024 & EgoVLP & Llama2-7B & 0.882 & 0.711 & 0.647 \\
    & ICVL~\cite{cao2025vision} & 2025 & CLIP & Llama3-8B & 0.857 & \textbf{0.652} & 0.619 \\
    & INSIGHT~\cite{chu2026intention} & 2026 & EgoVideo & Qwen2.5-VL-7B & \textbf{0.846} & 0.664 & \textbf{0.609} \\
    
    \midrule
    \multirow{10}{*}{\rotatebox{90}{Test}}
    & EgoT2~\cite{xue2023egocentric} & 2023 & SlowFast & -- & 0.935 & 0.722 & 0.764 \\
    & I-CVAE~\cite{mascaro2023intention} & 2023 & SlowFast & -- & 0.931 & 0.753 & 0.749 \\
    & HierVL~\cite{ashutosh2023hiervl} & 2023 & FrozenInTime & -- & 0.928 & 0.724 & 0.735 \\
    & ScalAnt~\cite{zhong2026scalable} & 2026 & VideoMAE & -- & 0.879 & 0.650 & 0.658 \\
    
    & VideoLLM~\cite{chen2023videollm} & 2023 & -- & GPT2 & 0.921 & 0.721 & 0.725 \\
    & AntGPT~\cite{zhao2024antgpt} & 2024 & EgoVLP & Llama2-7B & 0.877 & 0.650 & 0.650 \\
    & VideoLLM-O.~\cite{chen2024videollm} & 2024 & SigLIP & Llama3-8B & 0.889 & 0.691 & 0.692 \\
    & PALM~\cite{kim2024palm} & 2024 & EgoVLP & Llama2-7B & \textbf{0.850} & 0.647 & \textbf{0.612} \\
    & VideoLLM-M.~\cite{wu2024videollm} & 2024 & SigLIP & Llama3-8B & 0.884 & 0.689 & 0.676 \\
    & BiAnt~\cite{sato2025bidirectional} & 2025 & -- & Llama2-7B & 0.866 & \textbf{0.638} & 0.647 \\
    
    \bottomrule
    \end{tabular}} \vspace{-2mm}
\end{table}

\rev{
Table~\ref{tab:sota_ego4d} presents long-term action anticipation results on the large-scale egocentric Ego4D dataset~\cite{graumanEgo4DWorld0002022}, where methods observe 8 video segments and predict the next 20 actions. Performance is evaluated by edit distance (lower is better) over action, verb, and noun sequences, and reported on three splits: Val-v1, Val-v2 (approximately twice as large as Val-v1), and Test.
}
\rev{
Compared to the Breakfast and 50Salads benchmarks (Section~\ref{sec:breakfast_50salads}), LLMs are considerably more dominant on Ego4D. Specifically, on Val-v1, PlausiVL~\cite{mittal2024can} reduces both verb and noun edit distances by roughly 4–5 points over the best non-LLM method (HierVL~\cite{ashutosh2023hiervl}). On Val-v2, INSIGHT~\cite{chu2026intention} and ICVL~\cite{cao2025vision} push action edit distance below 0.86. On the test set, PALM~\cite{kim2024palm} and BiAnt~\cite{sato2025bidirectional} achieve the strongest results, substantially outperforming earlier non-LLM baselines such as EgoT2~\cite{xue2023egocentric} and HierVL. This likely reflects two factors: Ego4D is substantially larger, providing sufficient data for effective LLM fine-tuning, and unlike Breakfast, it does not require explicit duration prediction, a structured temporal reasoning task that may benefit from dedicated training mechanisms rather than the open-ended generative capabilities of LLMs.
}

\section{Challenges and Future Directions}
Despite the substantial advancements that have been achieved in this field, there is still potential for enhancing state-of-the-art algorithms. In the following discussion, we address the research challenges and suggest several promising directions for future research.

\textit{Expanding dataset coverage}:
Current anticipation methods are mainly trained and evaluated on datasets with restricted scene context and perspective, such as Breakfast~\cite{kuehneLanguageActionsRecovering2014} and EpicKitchens~\cite{damen2022rescaling}, limiting their ability to generalize to other scenarios. To make the anticipation methods applicable in our everyday life, more comprehensive datasets are needed that cover a wide range of daily life scenarios and include diverse actions across different cultures, geographies, and social contexts. While Ego4D~\cite{graumanEgo4DWorld0002022} is a strong example for diversity, it is limited to egocentric views. \rev{Ego-Exo4D~\cite{grauman2024ego} addresses this by providing simultaneously captured egocentric and exocentric video of skilled activities across 13 cities, opening new opportunities for studying anticipation from complementary viewpoints.} 
Studying anticipation across different tasks and geographic locations is also a promising direction, as recently explored for egocentric action recognition~\cite{plizzari2023generalisation}.           
It would also be valuable to include complex scenarios involving interactions with various objects or activities with long-term dependencies.
Since actions are influenced by their environment, contextual information such as location, time, or objects present could accompany dataset releases and be exploited by anticipation methods.
However, labeling natural, untrimmed, long videos is labor-intensive. Unsupervised approaches, not necessarily restricted to short-term anticipation, may thus present a promising direction. Furthermore, synthetic datasets generated from games~\cite{roitberg2021let} could also be of interest.

\textit{Exploiting language models}: 
Many actions follow a specific order (see Fig.~\ref{fig:predictability}). Identifying preceding actions could enable predicting subsequent ones without requiring frame-level feature prediction and classification. Moreover, humans adhere to daily routines, and learning such action patterns is akin to our predictive abilities based on lifelong experience. 
In this context, transferring the knowledge harnessed by large language models (LLMs) to action anticipation has emerged as an active research direction, as LLMs are trained on extensive text
corpora that encapsulate commonsense knowledge about action sequences and demonstrate remarkable zero-shot capability across multiple vision-language tasks.
\rev{
Recent works have begun to leverage LLMs for long-term action
anticipation, particularly on the Ego4D benchmark~\cite{graumanEgo4DWorld0002022,zhao2024antgpt,kim2024palm},
where promising results have been reported.
However, the application of LLMs to short-term anticipation settings, such as EpicKitchens~\cite{damen2022rescaling}, remains largely unexplored. Similarly, long-term anticipation benchmarks that additionally require the prediction of action durations, such as Breakfast~\cite{kuehneLanguageActionsRecovering2014},  have received limited attention from LLM-based methods. Bridging these gaps, namely adapting LLMs to fine-grained temporal reasoning and duration estimation, constitutes a promising direction for future work.
}

\textit{Personalization}: Current approaches are subject-agnostic, treating all individuals as having uniform preferences and behaviors. However, it is evident that each individual possesses unique tendencies, preferences, and patterns of behavior. Adapting anticipation models to individuals is another promising research direction. In particular, if the personal context, e.g., on vacation, is known. 

\textit{Addressing uncertainty with probabilistic generative models}: Generative models, including GANs~\cite{goodfellow2014gan}, VAEs~\cite{kingma2013auto}, Normalizing Flows~\cite{rezende2015variational}, and Diffusion Models~\cite{ho2020ddpm,liu2023diffact}, offer a promising avenue for addressing uncertainty in action anticipation. These models possess the capability to generate diverse potential future actions given a current context, thereby providing a richer understanding of the range of possible outcomes. 
\rev{
In particular, recent diffusion-based approaches have shown encouraging results on benchmarks such as Breakfast~\cite{kuehneLanguageActionsRecovering2014,zatsarynna2025manta}.
However, a critical limitation lies in the current evaluation protocols: existing metrics primarily assess whether the ground-truth future action label is contained among the generated predictions, rather than evaluating the overall quality and plausibility of the predicted sequences. This is inherently suboptimal, as current datasets provide only a single annotated future per observation rather than multiple equally valid continuations. 
Addressing this gap calls for new evaluation paradigms. One promising direction is to leverage LLMs as automated judges that assess the semantic plausibility and coherence of predicted action sequences, analogous to their growing use as evaluators in natural language generation~\cite{zheng2023judging}.
}
More broadly, methods for quantifying and incorporating predictive uncertainty into downstream decision-making processes remain an important open challenge.

\textit{Real-time action anticipation}: Although vision-based forecasting systems are typically designed for real-time deployment on autonomous entities such as self-driving cars and robots, they are predominantly evaluated in settings where inference latency is overlooked~\cite{li2020streaming,furnari2022streaming,FURNARI2023103763}. 
\rev{
This discrepancy is becoming increasingly relevant as the field adopts larger and more expressive
model components: modern visual encoders based on Vision Transformers~\cite{dosovitskiyImageWorth16x162021} and, more recently, large language models used for action reasoning, offer substantial gains in representational capacity but also introduce significant computational overhead. Designing anticipation methods that balance the predictive power of such large-scale models with the strict latency constraints of real-world applications is therefore an important research direction.
}

\textit{Multi-person action anticipation}: Existing methods mainly focus on single-person scenarios, but many real-world situations involve multiple people. Multi-person action forecasting in a video sequence has therefore surged as an intriguing topic. To address this, object and person detection can be incorporated, following the works of~\cite{sunRelationalActionForecasting2019,li2021restep}.

\section{Conclusion}
In this survey, we provide a systematic overview of over \rev{90} action anticipation methods pertinent to daily-living scenarios. We examined these methods from three perspectives: the specific research question each method addresses, a description of each method, and its contributions over previous work. 
We introduced a taxonomy to organize the different methods based on their main contributions. 
Moreover, we have provided a comparative summary of these methods in a tabular format to allow readers to identify low-level details at a glance. We have also introduced commonly used datasets and metrics, and presented results on several standard benchmarks. Lastly, we have offered insights in the form of future research directions. In conclusion, action anticipation is a captivating and relatively recent research topic that is garnering increasing attention within the community and holds value for numerous intelligent decision-making systems. While considerable progress has been made, there remains a vast scope for improvement in action anticipation using deep learning techniques.

\ifCLASSOPTIONcompsoc
\section*{Acknowledgments}
\else
\section*{Acknowledgment}
\fi

This work was supported by the JuBot project which was made possible by funding from the Carl-Zeiss-Foundation. Juergen Gall has been supported by the ERC Consolidator Grant FORHUE (101044724). 

\ifCLASSOPTIONcaptionsoff
\newpage
\fi

\bibliographystyle{IEEEtran}
\bibliography{reference}

\newlength{\mylength}
\setlength{\mylength}{0.8cm}

\vspace{-\mylength}
\begin{IEEEbiography}[{\includegraphics[width=1in,height=1.25in,clip,keepaspectratio]{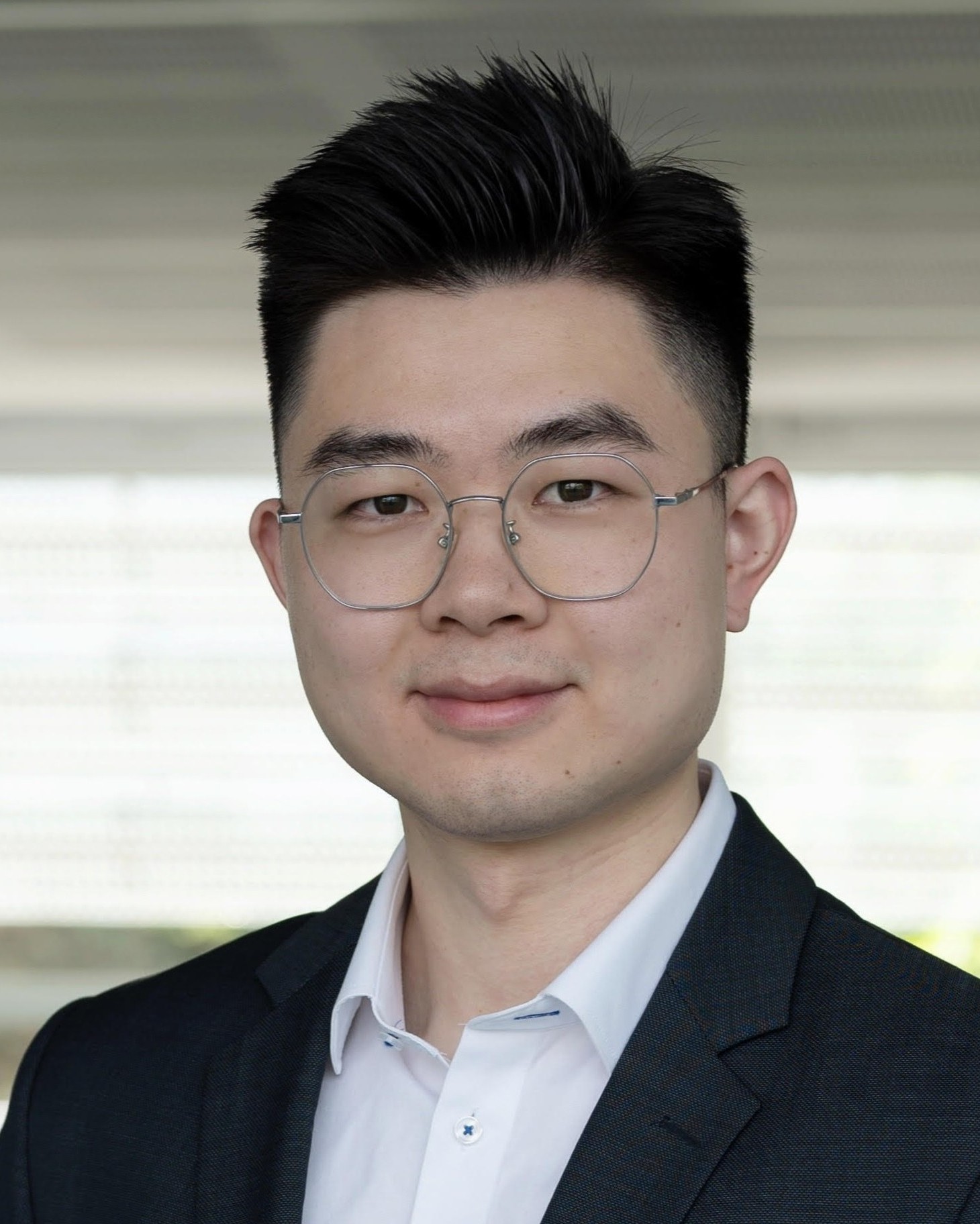}}]{Zeyun Zhong}
received the B.Eng. degree in process-, energy- and environmental engineering from Hannover University of Applied Sciences and Arts, in 2018, and the M.Sc. in mechatronics from Leibniz University Hannover, in 2021. He is currently working toward the Ph.D. degree at the Department of Informatics, Karlsruhe Institute of Technology. His main research interests include deep learning, video understanding, action recognition, and anticipation.
\end{IEEEbiography}

\vspace{-\mylength}
\begin{IEEEbiography}[{\includegraphics[width=1in,height=1.25in,clip,keepaspectratio]{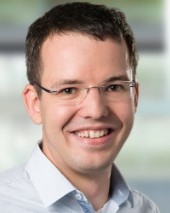}}]{Manuel Martin}
studied computer science at the Karlsruhe Institute of Technology (KIT) obtaining his diploma in 2013. He received his Ph.D. from KIT in 2023 for his work on 3D human body pose estimation and activity recognition of drivers in automated vehicles. He is now senior scientist in the group Perceptual User Interfaces at Fraunhofer IOSB, Karlsruhe, continuing his work on machine learning based occupant monitoring systems.
\end{IEEEbiography}

\vspace{-\mylength}
\begin{IEEEbiography}[{\includegraphics[width=1in,height=1.25in,clip,keepaspectratio]{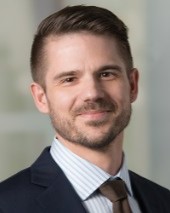}}]{Michael Voit}
studied computer science at the former University of Karlsruhe (TH), now Karlsruhe Institute of Technology (KIT), obtaining his diploma in 2004. He received his Ph.D. from KIT in 2012 for estimating the visual focus of attention of people by monitoring them with multiple overhead cameras distributed in a room. He is now head of the research group Perceptual User Interfaces at Fraunhofer IOSB, Karlsruhe, continuing his work on camera-based human machine interaction techniques.
\end{IEEEbiography}

\vspace{-\mylength}
\begin{IEEEbiography}[{\includegraphics[width=1in,height=1.25in,clip,keepaspectratio]{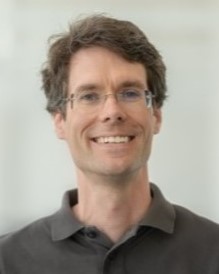}}]{Juergen Gall} obtained his B.Sc. and his Masters degree in mathematics from the University of Wales Swansea (2004) and from the University of Mannheim (2005). In 2009, he obtained a
Ph.D.\ in computer science from the Saarland University and the Max Planck Institut f\"ur Informatik. He was a postdoctoral researcher at the Computer Vision Laboratory, ETH Zurich, from 2009 until 2012 and senior research scientist at the Max Planck Institute for Intelligent Systems in T\"ubingen from 2012 until 2013. Since 2013, he is professor at the University of Bonn and member of the Lamarr Institute for Machine Learning and Artificial Intelligence.
\end{IEEEbiography}

\vspace{-\mylength}
\begin{IEEEbiography}[{\includegraphics[width=1in,height=1.25in,clip,keepaspectratio]{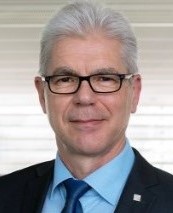}}]{Jürgen Beyerer}
studied electrical engineering at the University of Karlsruhe, obtaining his diploma in 1989. In 1994, he obtained his Ph.D. from the Institute for Measurement and Control Systems (MRT) at the same university. Post his doctorate, he established a research group on automatic visual inspection and image processing at MRT. From 1999 to 2004, he led the Hottinger Systems GmbH in Mannheim and served as the deputy managing director for Hottinger Maschinenbau GmbH. Since 2004, he is professor at the Karlsruhe Institute of Technology and concurrently the head of the Fraunhofer Institute of Optronics, System Technologies and Image Exploitation (IOSB).
\end{IEEEbiography}

\vfill

\clearpage
\section{Supplementary Material}
\balance

While Tables \ref{tab:tvseries_thumos14}-\ref{tab:breakfast_50salads} show the results for a subset of the evaluation protocols, we provide the comprehensive results in Tables \ref{tab:tvseries}-\ref{tab:50salads}. 
These results encompass both short-term and long-term action anticipation, covering both third-person and egocentric perspectives.

\begin{table*}[b!]
\centering
\caption{Comparison of the state-of-the-art methods on TVSeries~\cite{geest2016online} in terms of mean cAP (\%). The optimal performance in each column within every block is indicated in bold, while the supreme overall performance in each column is further denoted by an underscore.}
\begin{tabular}{@{}lccccccccccccc@{}}
\toprule
\multirow{2}{*}{Method} & \multirow{2}{*}{Year} & \multirow{2}{*}{Backbone} & \multirow{2}{*}{Pre-train} & \multirow{2}{*}{\makecell{Obs. \\ len.}} & \multicolumn{8}{c}{mcAP@Time predicted into the future (seconds)} \\ \cmidrule(lr){6-13}
 & & & & & 0.25 & 0.5 & 0.75 & 1.0 & 1.25 & 1.5 & 1.75 & 2.0 & Avg. \\
\midrule
ED~\cite{gaoREDReinforcedEncoderDecoder2017} & 2017 & \multirow{3}{*}{VGG} & \multirow{3}{*}{IN1K} & 4s & 71.0 & 70.6 & 69.9 & 68.8 & 68.0 & 67.4 & 67.0 & 66.7 & 68.7\\
RED~\cite{gaoREDReinforcedEncoderDecoder2017} & 2017 & & & 4s & 71.2 & 71.0 & 70.6 & 70.2 & 69.2 & 68.5 & 67.5 & 66.8 & 69.4 \\
TTPP\cite{wangTTPPTemporalTransformer2020} & 2020 & & & 2s & \textbf{72.7} & \textbf{72.3} & \textbf{71.9} & \textbf{71.6} & \textbf{71.3} & \textbf{70.9} & \textbf{69.9} & \textbf{69.3} & \textbf{71.3} \\
\midrule
ED~\cite{gaoREDReinforcedEncoderDecoder2017} & 2017 & \multirow{9}{*}{TS} & \multirow{9}{*}{ANet1.3} & 4s & 78.5 & 78.0 & 76.3 & 74.6 & 73.7 & 72.7 & 71.7 & 71.0 & 74.5 \\
RED~\cite{gaoREDReinforcedEncoderDecoder2017} & 2017 & & & 4s & 79.2 & 78.7 & 77.1 & 75.5 & 74.2 & 73.0 & 72.0 & 71.2 & 75.1 \\
TRN~\cite{xu2019trn} & 2019 & & & 16s & 79.9 & 78.4 & 77.1 & 75.9 & 74.9 & 73.9 & 73.0 & 72.3 & 75.7 \\
TTPP\cite{wangTTPPTemporalTransformer2020} & 2020 & & & 2s & 81.2 & 80.3 & 79.3 & 77.6 & 76.9 & 76.7 & 76.0 & 74.9 & 77.9\\
LAP~\cite{qu2020lapnet}  & 2020 & & & 16s & \textbf{82.6} & \textbf{81.3} & \textbf{80.0} & \textbf{78.9} & \textbf{77.9} & \textbf{77.1} & \textbf{76.3} & \textbf{75.5} & 78.7 \\
OadTR\cite{wang2021oadtr} & 2021 & & & 63s & 81.9 & 80.6 & 79.4 & 78.2 & 77.1 & 76.0 & 75.2 & 74.3 & 77.8 \\
LSTR~\cite{xu2021lstr}  & 2021 & & & 520s & -- & -- & -- & -- & -- & -- & -- & -- & 80.8 \\
\rev{MAT}~\cite{wang2023memory} & 2023 & & & 264s & -- & -- & -- & -- & -- & -- & -- & -- & 81.5 \\ 
\rev{BiOMamba}~\cite{wang2025biomamba} & 2025 & & & 264s & -- & -- & -- & -- & -- & -- & -- & -- & \textbf{82.9} \\ 
\midrule
OadTR\cite{wang2021oadtr} & 2021 & \multirow{3}{*}{TS} & \multirow{3}{*}{K400} & 63s & \textbf{\ul{84.1}} & \textbf{\ul{82.6}} & \textbf{\ul{81.3}} & \textbf{\ul{80.1}} & \textbf{\ul{78.9}} & \textbf{\ul{77.7}} & \textbf{\ul{76.7}} & \textbf{\ul{75.7}} & 79.1 \\
\rev{MAT}~\cite{wang2023memory} & 2023 & & & 264s & -- & -- & -- & -- & -- & -- & -- & -- & 82.6 \\ 
\rev{BiOMamba}~\cite{wang2025biomamba} & 2025 & & & 264s & -- & -- & -- & -- & -- & -- & -- & -- & \textbf{\ul{83.7}} \\ 
\bottomrule
\end{tabular}
\label{tab:tvseries}
\end{table*}

\noindent\textbf{TVSeries}~\cite{geest2016online} \textbf{\& THUMOS14}~\cite{thumos2014}. TVSeries contains six popular TV series including 30 realistic, everyday actions. THUMOS14 contains sport videos depicting 20 actions. Both datasets employ an anticipation protocol to predict actions at various future timestamps, ranging from 0.25 to 2.0 seconds. For evaluation, mean calibrated average precision (mcAP) is used for TVSeries and mean average precision (mAP) for THUMOS14. Comprehensive benchmark results and average performance across all timestamps can be found in Table~\ref{tab:tvseries} and Table~\ref{tab:thumos14}, respectively.

\noindent\textbf{EpicKitchens-100}~\cite{damen2022rescaling} consists of long unscripted videos of cooking activities
totalling 100 hours. Table~\ref{tab:ek100_sota} outlines
three performance benchmarks: \textit{overall}, \textit{unseen kitchen}, and \textit{tail classes}. Each of these benchmarks is evaluated based on the top-5 recall for verbs, nouns, and actions across all
classes. The primary metric used to rank the methods is accentuated. If the anticipation approach employs a backbone fine-tuned for feature extraction, it is marked with a check mark in the column titled \textit{E2E}.

\noindent\textbf{Breakfast}~\cite{kuehneLanguageActionsRecovering2014} \textbf{\& 50Salads}~\cite{stein2013combining}. The Breakfast dataset features videos of various individuals preparing breakfast in distinct kitchens, while the 50Salads dataset consists of top-view RGB-D video footage capturing individuals making salads. Evaluation of these datasets is conducted using mean-over-classes accuracy, averaged across future timestamps within a specified anticipation duration. For evaluation, a fraction ($\alpha$) of a complete video is observed, with the objective of predicting actions in the subsequent video. Typically, the observation ratio $\alpha$ is set to 0.2 or 0.3. Action predictions are then made in $\beta$ segments of the full video, where $\beta$ can be $\{ 0.1, 0.2, 0.3, 0.5\}$. 

\begin{table*}[t]
\centering
\caption{Comparison of the state-of-the-art methods on THUMOS14~\cite{thumos2014} in terms of mAP (\%). For details on the font coding, please refer to Table~\ref{tab:tvseries}.}
\begin{tabular}{@{}lcccccccccccccc@{}}
\toprule
\multirow{2}{*}{Method} & \multirow{2}{*}{Year} & \multirow{2}{*}{Backbone} & \multirow{2}{*}{Pre-train} & \multirow{2}{*}{\makecell{Obs. \\ len.}} & \multicolumn{8}{c}{mAP@Time predicted into the future (seconds)} \\ \cmidrule(lr){6-13}
 & & & & & 0.25 & 0.5 & 0.75 & 1.0 & 1.25 & 1.5 & 1.75 & 2.0 & Avg. \\
\midrule
ED~\cite{gaoREDReinforcedEncoderDecoder2017} & 2017 & \multirow{11}{*}{TS} & \multirow{11}{*}{ANet1.3} & 4s & 43.8 & 40.9 & 38.7 & 36.8 & 34.6 & 33.9 & 32.5 & 31.6 & 36.6\\
RED~\cite{gaoREDReinforcedEncoderDecoder2017} & 2017 & & & 4s & 45.3 & 42.1 & 39.6 & 37.5 & 35.8 & 34.4 & 33.2 & 32.1 & 37.5 \\
TRN~\cite{xu2019trn} & 2019 & & & 16s &45.1 & 42.4 & 40.7 & 39.1 & 37.7 & 36.4 & 35.3 & 34.3 & 38.9 \\
TTPP~\cite{wangTTPPTemporalTransformer2020} & 2020 & & & 2s &45.9 & 43.7 & 42.4 & 41.0 & 39.9 & 39.4 & 37.9 & 37.3 & 40.9 \\
LAP~\cite{qu2020lapnet} & 2020 & & & 16s &49.0 & 47.4 & 45.3 & 43.2 & 41.3 & 39.7 & 38.3 & 37.0 & 42.6 \\
OadTR~\cite{wang2021oadtr} & 2021 & & & 63s &50.2 & 49.3 & 48.1 & 46.8 & 45.3 & 43.9 & 42.4 & 41.1 & 45.9 \\
LSTR~\cite{xu2021lstr} & 2021 & & & 520s &-- & -- & -- & -- & -- & -- & -- & -- & 50.1 \\
TeSTra~\cite{zhao2022realtime} & 2022 & &  & 520s &64.7 & 61.8 & 58.7 & 55.7 & 53.2 & 51.1 & 49.2 & 47.8 & 55.3\\
\rev{MAT}~\cite{wang2023memory} & 2023 & & & 264s &-- & -- & -- & -- & -- & -- & -- & -- & 57.3 \\ 
\rev{MiniROAD}~\cite{an2023miniroad} & 2023 & & & -- &65.4 & 63.3 & 60.5 & 57.4 & 54.8 & 51.9 & 49.7 & 47.7 & 56.3 \\
\rev{BiOMamba}~\cite{wang2025biomamba} & 2025 & & & 264s &\textbf{68.3} & \textbf{65.1} & \textbf{62.0} & \textbf{58.9} & \textbf{56.4} & \textbf{54.5} & \textbf{52.2} & \textbf{50.6} & \textbf{58.5} \\ 

\midrule

TTPP~\cite{wangTTPPTemporalTransformer2020}& 2020 & \multirow{9}{*}{TS} & \multirow{9}{*}{K400} & 2s & 46.8 & 45.5 & 44.6 & 43.6 & 41.9 & 41.1 & 40.4 & 38.7 & 42.8 \\
LSTR~\cite{xu2021lstr} & 2021 & & & 520s &60.4 & 58.6 & 56.0 & 53.3 & 50.9 & 48.9 & 47.1 & 45.7 & 52.6\\
OadTR~\cite{wang2021oadtr} & 2021 & & & 63s &59.8 & 58.5 & 56.6 & 54.6 & 52.6 & 50.5 & 48.6 & 46.8 & 53.5 \\
TeSTra~\cite{zhao2022realtime} & 2022 & & & 520s &66.2 & 63.5 & 60.5 & 57.4 & 54.8 & 52.6 & 50.5 & 48.9 & 56.8\\
\rev{MAT}~\cite{wang2023memory} & 2023 & & & 264s &-- & -- & -- & -- & -- & -- & -- & -- & 58.2 \\ 
\rev{MiniROAD}~\cite{an2023miniroad} & 2023 & & & -- &68.5 & 66.3 & 63.5 & 60.7 & 57.9 & 55.6 & 53.5 & 51.4 & 59.7 \\
\rev{JOADAA}~\cite{guermal2024joadaa} & 2024 & & & 136s &67.7 & 63.9 & -- & \textbf{\ul{62.9}} & -- & \textbf{\ul{59.3}} & -- & -- & -- \\
\rev{BiOMamba}~\cite{wang2025biomamba} & 2025 & & & 264s &70.0 & 66.7 & 63.4 & 60.2 & 57.6 & 55.3 & 53.2 & 51.5 & 59.7 \\ 
\rev{ScalAnt}~\cite{zhong2026scalable} & 2026 & & & -- &\textbf{\ul{71.7}} & \textbf{\ul{69.3}} & \textbf{\ul{66.2}} & 62.5 & \textbf{\ul{59.6}} & 56.8 & \textbf{\ul{54.5}} & \textbf{\ul{52.5}} & \textbf{\ul{61.6}} \\ 
\bottomrule
\end{tabular}

\label{tab:thumos14}
\end{table*}

\begin{table*}[t]
    \centering
    \setlength\tabcolsep{4pt}
    \caption{Comparison of state-of-the-art methods on the validation and test set of EpicKitchens-100 in terms of mean top-5 recall (\%). The optimal performance in each column within every block is indicated in bold. Modalities: Objects (O), Bounding Boxes (BB), Motion (M), Audio (Au).}
    \resizebox{\linewidth}{!}{
    \begin{tabular}{@{}clclcccccc>{\columncolor{Gray}}c>{\enspace}ccc>{\enspace}ccc@{}}
        \toprule
         \multirow{2}{*}{} & \multirow{2}{*}{Method} & \multirow{2}{*}{Year} & \multirow{2}{*}{Modality} & \multirow{2}{*}{Backbone} & \multirow{2}{*}{Init.} & \multirow{2}{*}{E2E} & \multirow{2}{*}{\makecell{Obs. \\ len.}}  & \multicolumn{3}{c}{Overall} & \multicolumn{3}{c}{Unseen Kitchen} & \multicolumn{3}{c}{Tail Classes}\\ 
          \cmidrule(lr){9-11} \cmidrule(lr){12-14} \cmidrule(lr){15-17}
          & & & & & & & & Verb & Noun & Act. & Verb & Noun & Act. & Verb & Noun & Act. \\ \midrule
         
         \multirow{27}{*}{\rotatebox{90}{Val}} 
         & RULSTM~\cite{furnariWhatWouldYou2019} & 2019 & RGB & TSN & IN1K & \xmark & 2.75s  & 27.5 & 29.0 & 13.3 & -- & -- & -- & -- & -- & -- \\
         & TempAgg \cite{senerTemporalAggregateRepresentations2020} & 2020 & RGB & TSN & IN1K & \xmark & 6s & 24.2 & 29.8 & 13.0 & 27.0 & 23.0 & 12.2 & 16.2 & 22.9 & 10.4\\
         & AVT~\cite{girdharAnticipativeVideoTransformer2021} & 2021 & RGB & TSN & IN1K & \xmark & 10s  & 27.2 & 30.7 & 13.6  & -- & -- & -- & -- & -- & --   \\
         & AVT~\cite{girdharAnticipativeVideoTransformer2021} & 2021 & RGB & ViT-B & IN21K & \checkmark & 10s & 30.2 & 31.7 & 14.9  & -- & -- & -- & -- & -- & --   \\
         & MeMViT \cite{wuMeMViTMemoryAugmentedMultiscale2022} & 2022 & RGB & MViT-B & K400 & \checkmark & 70.4s & 32.8 & 33.2 & 15.1 & 27.5 & 21.7 & 9.8 & 26.3 & 27.4 & 13.2 \\
         & MeMViT \cite{wuMeMViTMemoryAugmentedMultiscale2022} & 2022 & RGB & MViT-L & K700 & \checkmark & 70.4s & 32.2 & 37.0 & 17.7 & 28.6 & 27.4 & 15.2 & 25.3 & 31.0 & 15.5 \\
         & DCR~\cite{xuLearningAnticipateFuture2022} & 2022 & RGB & TSM & K400 & \xmark & 10s & 32.6 & 32.7 & 16.1 & -- & -- & -- & -- & -- & -- \\
         & RAFTformer~\cite{girase2023latency} & 2023 & RGB & MViT-B & K400 + IN1K & \xmark & 16s & 33.3 & 35.5 & 17.6 & -- & -- & -- & -- & -- & -- \\
         & RAFTformer~\cite{girase2023latency} & 2023 & RGB & MViT-B & K700 & \xmark & 16s & 33.7 & 37.1 & 18.0 & -- & -- & -- & -- & -- & -- \\

         & \rev{S-GEAR}~\cite{diko2024semantically} & 2024 & RGB & ViT-B & IN21K & \checkmark & 15s & 31.1 & 37.3 & 18.3 & -- & -- & -- & -- & -- & -- \\ 
         & \rev{ARR}~\cite{liu2024recognition} & 2024 & RGB & AIM & -- & \checkmark & 8s & 31.3 & 32.0 & 16.3 & -- & -- & -- & -- & -- & -- \\ 
         & \rev{CPM}~\cite{xie2024towards} & 2024 & RGB & TSM & K400 & \xmark & 30f & -- & -- & 17.2 & -- & -- & -- & -- & -- & -- \\ 
         & \rev{UADT}~\cite{guo2024uncertainty} & 2024 & RGB & MViT-L & K700 & \xmark & -- & 38.2 & 41.4 & 20.3 & -- & -- & -- & -- & -- & -- \\ 
         & \rev{PlausiVL}~\cite{mittal2024can} & 2024 & RGB & ViT-G/14 & -- & \checkmark & 32f & \textbf{55.6} & \textbf{54.2} & \textbf{27.6} & \textbf{49.5} & \textbf{53.9} & \textbf{27.0} & \textbf{48.4} & \textbf{41.3} & \textbf{22.1} \\ 
         & \rev{ScalAnt}~\cite{zhong2026scalable} & 2026 & RGB & TSN & IN1K & \xmark & 32s & 31.1 & 27.2 & 17.6 & -- & -- & -- & -- & -- & -- \\ 
         & \rev{AGA}~\cite{tai2026action} & 2026 & RGB & TSN & IN1K & \xmark & 30s & 32.2 & 35.7 & 17.5 & 31.4 & 24.9 & 11.9 & 26.9 & 31.8 & 16.8 \\ 
         & \rev{AGA}~\cite{tai2026action} & 2026 & RGB & Swin-B & K400 & \xmark & 30s & 32.5 & 38.7 & 18.8 & 34.4 & 28.5 & 16.3 & 27.4 & 35.0 & 18.4 \\ 

         \arrayrulecolor{lightgray}\hhline{*{1}{~}*{14}{-}}\arrayrulecolor{black} \rule{0pt}{2.6ex}
         
         & RULSTM \cite{furnariWhatWouldYou2019} & 2019 & RGB,O,M & TSN & IN1K & \xmark & 2.75s & {27.8} & 30.8 & 14.0 & {28.8} & \textbf{27.2} & 14.2 & {19.8} & 22.0 & 11.1 \\
         & TempAgg~\cite{senerTemporalAggregateRepresentations2020} & 2020 & RGB,O,M,BB & TSN & IN1K & \xmark & 6 & 23.2 & 31.4 & 14.7 & 28.0 & 26.2 & 14.5 & 14.5 & 22.5 & 11.8 \\
         & AVT~\cite{girdharAnticipativeVideoTransformer2021} & 2021 & RGB,O & ViT-B & IN21K & \checkmark & 10s & 28.2 & 32.0 & 15.9 & 29.5 & 23.9 & 11.9 & 21.1 & 25.8 & 14.1 \\
         & DCR~\cite{xuLearningAnticipateFuture2022} & 2022 & RGB,O & TSM & K400 & \xmark & 10s & -- & -- & 18.3 & -- & -- & 14.7 & -- & -- & 15.8 \\
         & AFFT~\cite{zhong2023afft} & 2023 & RGB,O,M & TSN & IN1K & \xmark & 18s & 21.3 & 32.7 & 16.4 & 24.1 & 25.5 & 13.6 & 13.2 & 25.8 & 14.3 \\
         & AFFT~\cite{zhong2023afft} & 2023 & RGB,O,M,Au & Swin-B & K400 & \xmark & 16s & 22.8 & {34.6} & 18.5 & 24.8 & 26.4 & 15.5 & 15.0 & {27.7} & 16.2\\

         & \rev{InAViT}~\cite{roy2024interaction} & 2024 & RGB,HOI & MotionFormer & -- & \checkmark & 2s & \textbf{52.5} & \textbf{51.9} & \textbf{25.9} & \textbf{46.5} & \textbf{51.3} & \textbf{25.3} & \textbf{45.3} & \textbf{39.2} & \textbf{20.2} \\ 
         & \rev{S-GEAR}~\cite{diko2024semantically} & 2024 & RGB,O & ViT-B & IN21K & \checkmark & 15s  & 29.5 & 37.8 & 18.9 & -- & -- & -- & -- & -- & -- \\ 
         & \rev{UADT}~\cite{guo2024uncertainty} & 2024 & RGB,O,M & MViT-L & K700 & \xmark & -- & 43.5 & 46.6 & 23.0 & -- & -- & -- & -- & -- & -- \\ 
         & \rev{CPM}~\cite{xie2024towards} & 2024 & RGB,O,M & TSM & K400 & \xmark & 30f & -- & -- & 19.4 & -- & -- & -- & -- & -- & -- \\ 

         \midrule
         
         \multirow{8}{*}{\rotatebox{90}{Test}} 
         & RULSTM \cite{furnariWhatWouldYou2019} & 2019 & RGB,O,M & TSN & IN1K & \xmark & 2.75s & {25.3} & 26.7 & 11.2 & 19.4 & 26.9 & 9.7 & {17.6} & 16.0 & 7.9 \\
         & TempAgg \cite{senerTemporalAggregateRepresentations2020} & 2020 & RGB,O,M,BB & TSN & IN1K & \xmark & 6s  & 21.8 & 30.6 & 12.6 & 17.9 & 27.0 & 10.5 & 13.6 & 20.6 & 8.9 \\
         & AVT \cite{girdharAnticipativeVideoTransformer2021} & 2021 & RGB,O & ViT-B & IN21K & \checkmark & 10s & 25.6 & 28.8 & 12.6 & 20.9 & 22.3 & 8.8 & 19.0 & 22.0 & 10.1 \\
         & TCN-TBN \cite{zatsarynnaMultiModalTemporalConvolutional2021} & 2021 & RGB,O,M & TBN & IN1K & \xmark & 5.25s & 21.5 & 26.8 & 11.0 & {20.8} & 28.3 & 12.2 & 13.2 & 15.4 & 7.2 \\
         & Abst. goal~\cite{roy2022predicting} & 2022 & RGB,O,M & TSN & IN1K & \xmark & 2s & 31.4 & 30.1 & 14.3 & 31.4 & 35.6 & 17.3 & -- & -- & -- \\
         & AFFT~\cite{zhong2023afft} & 2023 & RGB,O,M,Au & Swin-B & K400 & \xmark & 16s & 20.7 &{31.8} & 14.9 &16.2 &27.7 &12.1 &13.4 & 23.8 & 11.8 \\
         & RAFTformer~\cite{girase2023latency} & 2023 & RGB & MViT-B & K400 + IN1K & \xmark & 16s & 27.3 & 32.8 & 14.0 & -- & -- & -- & -- & -- & -- \\
         & RAFTformer~\cite{girase2023latency} & 2023 & RGB & MViT-B & K700 & \xmark & 16s & 27.4 & 34.0 & 14.7 & -- & -- & -- & -- & -- & -- \\

         & \rev{InAViT}~\cite{roy2024interaction} & 2024 & RGB,HOI & MotionFormer & -- & \checkmark & 2s & \textbf{49.1} & \textbf{50.0} & \textbf{23.8} & \textbf{44.4} & \textbf{49.3} & \textbf{23.5} & \textbf{43.2} & \textbf{39.9} & \textbf{18.1} \\ 
         
         & \rev{S-GEAR}~\cite{diko2024semantically} & 2024 & RGB,O & ViT-B & IN21K & \checkmark & 15s & 25.9 & 32.0 & 14.7  & -- & -- & -- & -- & -- & -- \\
         
         & \rev{AGA}~\cite{tai2026action} & 2026 & RGB & Swin-B & K400 & \xmark & 30s & 30.8 & 36.4 & 16.9 & 22.3 & 30.0 & 13.5 & 25.8 & 30.0 & 14.9 \\ 

         \bottomrule
    \end{tabular}}
    \label{tab:ek100_sota}
\end{table*}

\begin{table*}[t]
    \centering
    \caption{Benchmark of long-term action anticipation on Breakfast~\cite{kuehneLanguageActionsRecovering2014} in terms of mean over classes (\%). For details on the font coding, please refer to Table~\ref{tab:tvseries}.}
    \begin{tabular}{@{}lcllcccccccc@{}}
    \toprule
    \multirow{2}{*}{Input Type}& \multirow{2}{*}{Backbone} & \multirow{2}{*}{Methods} & \multirow{2}{*}{Year} & \multicolumn{4}{c}{$\beta$ ($\alpha$ = 0.2)} & \multicolumn{4}{c}{$\beta$ ($\alpha$ = 0.3)} \\
    \cmidrule(lr){5-8} \cmidrule(lr){9-12}
    & & & & 0.1 & 0.2 & 0.3 & 0.5 & 0.1 & 0.2 & 0.3 & 0.5 \\
    \midrule
    \multirow{8}{*}{GT. label} & \multirow{8}{*}{--} & RNN~\cite{farhaWhenWillYou2018} & 2018 & 60.35 & 50.44 & 45.28 & 40.42 & 61.45 & 50.25 & 44.90 & 41.75 \\
    & & CNN~\cite{farhaWhenWillYou2018} & 2018 & 57.97 & 49.12 & 44.03 & 39.26 & 60.32 & 50.14 & 45.18 & 40.51\\
    & & Time-Cond.~\cite{keTimeConditionedActionAnticipation2019} & 2019 & 64.46 & 56.27 & 50.15 & 43.99 & 65.95 & 55.94 & 49.14 & 44.23 \\
    & & UAAA~\cite{abufarhaUncertaintyAwareAnticipationActivities2019} (avg.) & 2019 & 50.39 & 41.71 & 37.79 & 32.78 & 51.25 & 42.94 & 38.33 & 33.07 \\
    & & UAAA~\cite{abufarhaUncertaintyAwareAnticipationActivities2019} (top-1) & 2019 & 78.84 & \textbf{\ul{72.84}} & 66.29 & 63.45 & 82.00 & 72.83 & 69.13 & 62.39 \\
    & & TempAgg~\cite{senerTemporalAggregateRepresentations2020} & 2020 & 65.50 & 55.50 & 46.80 & 40.10 & 67.40 & 56.10 & 47.40 & 41.50 \\
    & & Zhao\etal~\cite{zhao2020diverse} (avg.) & 2020 & 72.22 & 62.40 & 56.22 & 45.95 & 74.14 & 71.32 & 65.30 & 52.38 \\
     & & Zhao\etal~\cite{zhao2020diverse} (top-1) & 2020 & \textbf{\ul{82.08}} & 70.59 & \textbf{\ul{68.51}} & \textbf{\ul{64.06}} &  \textbf{\ul{83.36}} & \textbf{\ul{76.85}} & \textbf{\ul{72.13}} & \textbf{\ul{64.06}} \\
    
    \midrule
     
    \multirow{7}{*}{Pred. label} & \multirow{6}{*}{Fisher} & RNN~\cite{farhaWhenWillYou2018} & 2018 & 18.11 & 17.20 & 15.94 & 15.81 & 21.64 & 20.02 & 19.73 & 19.21 \\
    & & CNN~\cite{farhaWhenWillYou2018} & 2018 & 17.90 & 16.35 & 15.37 & 14.54 & 22.44 & 20.12 & 19.69 & 18.76 \\
    & & UAAA~\cite{abufarhaUncertaintyAwareAnticipationActivities2019} (avg.) & 2019 & 15.69 & 14.00 & 13.30 & 12.95 & 19.14 & 17.18 & 17.38 & 14.98 \\
    & & UAAA~\cite{abufarhaUncertaintyAwareAnticipationActivities2019} (top-1) & 2019 & 28.89 & 28.43 & 27.61 & \textbf{28.04} & 32.38 & 31.60 & \textbf{32.83} & \textbf{30.79} \\
    & & Time-Cond.~\cite{keTimeConditionedActionAnticipation2019} & 2019 & 18.41 & 17.21 & 16.42 & 15.84 & 22.75 & 20.44 & 19.64 & 19.75 \\ 
    & & TempAgg~\cite{senerTemporalAggregateRepresentations2020} & 2020 & 18.80 & 16.90 & 16.50 & 15.40 & 23.00 & 20.00 & 19.90 & 18.60 \\[.8mm]

    \arrayrulecolor{lightgray}\hhline{*{1}{~}*{11}{-}}\arrayrulecolor{black} \rule{0pt}{2.6ex}
     & {I3D} & TempAgg~\cite{senerTemporalAggregateRepresentations2020} & 2020 & \textbf{37.40} & \textbf{31.20} & \textbf{30.00} & 26.10 & \textbf{39.50} & \textbf{34.10} & 31.00 & 27.90 \\

     \midrule
     
     \multirow{15}{*}{Features} & \multirow{2}{*}{Fisher}  & CNN~\cite{farhaWhenWillYou2018} & 2018 & 12.78 & 11.62 & 11.21 & 10.27 & 17.72 & 16.87 & 15.48 & 14.09 \\
    & & TempAgg~\cite{senerTemporalAggregateRepresentations2020} & 2020 & 15.60 & 13.10 & 12.10 & 11.10 & 19.50 & 17.00 & 15.60 & 15.10 \\[.8mm]
     
    \arrayrulecolor{lightgray}\hhline{*{1}{~}*{11}{-}}\arrayrulecolor{black} \rule{0pt}{2.6ex}
    & \multirow{13}{*}{I3D} & TempAgg~\cite{senerTemporalAggregateRepresentations2020} & 2020 & 24.20 & 21.10 & 20.00 & 18.10 & 30.40 & 26.30 & 23.80 & 21.20 \\
    & & Cycle Cons\cite{abu2020long} & 2020 & 25.88 & 23.42 & 22.42 & 21.54 & 29.66 & 27.37 & 25.58 & 25.20 \\
    & & A-ACT~\cite{gupta2022act} & 2022 & 26.70 & 24.30 & 23.20 & 21.70 & 30.80 & 28.30 & 26.10 & 25.80 \\
    & & FUTR~\cite{gongFutureTransformerLongterm2022} & 2022 & 27.70 & 24.55 & 22.83 & 22.04 & 32.27 & 29.88 & 27.49 & 25.87 \\

    & & \rev{DiffAnt}~\cite{zhong2023diffant} & 2023 & 25.33 & 24.59 & 24.39 & 22.74 & 32.13 & 31.83 & 31.18 & 30.77 \\ 
    & & \rev{ActFusion}~\cite{gong2024actfusion} & 2024 & 28.25 & 25.52 & 24.66 & 23.25 & 35.79 & 31.76 & 29.64 & 28.78 \\ 
    & & \rev{GTD}~\cite{zatsarynna2024gated} (avg.) & 2024 & 24.00 & 22.00 & 21.40 & 20.60 & 29.10 & 26.80 & 25.30 & 24.20 \\ 
    & & \rev{GTD}~\cite{zatsarynna2024gated} (top-1) & 2024 & 51.20 & 47.30 & 45.60 & 45.00 & 54.00 & 50.40 & 49.60 & 47.80 \\ 
    & & \rev{ActionLLM}~\cite{wang2025multimodal} & 2025 & 26.38 & 23.32 & 21.70 & 20.43 & 31.77 & 29.45 & 26.68 & 24.55 \\ 
    & & \rev{MANTA}~\cite{zatsarynna2025manta} (avg.) & 2025 & 27.70 & 25.30 & 24.60 & 23.80 & 34.20 & 30.90 & 29.10 & 27.70 \\ 
    & & \rev{MANTA}~\cite{zatsarynna2025manta} (top-1) & 2025 & {55.50} & {51.00} & {47.90} & {46.90} & 59.60 & 55.00 & 53.70 & 51.90 \\ 
    & & \rev{MixANT}~\cite{wasim2025mixant} (avg.) & 2025 & 29.60 & 26.30 & 25.90 & 25.00 & 36.20 & 32.80 & 31.20 & 28.70 \\ 
    & & \rev{MixANT}~\cite{wasim2025mixant} (top-1) & 2025 & \textbf{57.10} & \textbf{52.00} & \textbf{49.10} & \textbf{48.40} & \textbf{60.70} & \textbf{56.30} & \textbf{55.50} & \textbf{53.50} \\ 
     
     \bottomrule
    \end{tabular}
    \label{tab:breakfast}
\end{table*}

\begin{table*}[t]
    \centering
    \caption{Benchmark of long-term action anticipation on 50 Salads~\cite{stein2013combining} in terms of mean over classes (\%). For details on the font coding, please refer to Table~\ref{tab:tvseries}.}
    \begin{tabular}{@{}lcllcccccccc@{}}
    \toprule
    \multirow{2}{*}{Input Type}& \multirow{2}{*}{Backbone} & \multirow{2}{*}{Methods} & \multirow{2}{*}{Year} & \multicolumn{4}{c}{$\beta$ ($\alpha$ = 0.2)} & \multicolumn{4}{c}{$\beta$ ($\alpha$ = 0.3)} \\
    \cmidrule(lr){5-8} \cmidrule(lr){9-12}
     & & & & 0.1 & 0.2 & 0.3 & 0.5 & 0.1 & 0.2 & 0.3 & 0.5 \\
    \midrule
    \multirow{8}{*}{GT. label} & \multirow{8}{*}{--} & RNN~\cite{farhaWhenWillYou2018} & 2018 & 42.30 & 31.19 & 25.22 & 16.82 & 44.19 & 29.51 & 19.96 & 10.38 \\
     & & CNN~\cite{farhaWhenWillYou2018} & 2018 & 36.08 & 27.62 & 21.43 & 15.48 & 37.36 & 24.78 & 20.78 & 14.05 \\
     & & Time-Cond.~\cite{keTimeConditionedActionAnticipation2019} & 2019 & 45.12 & 33.23 & 27.59 & 17.27 & 46.40 & 34.80 & 25.24 & 13.84 \\
     & & UAAA~\cite{abufarhaUncertaintyAwareAnticipationActivities2019} (avg.) & 2019 & 34.95 & 28.05 & 24.08 & 15.41 & 33.15 & 24.65 & 18.84 & 14.34 \\
     & & UAAA~\cite{abufarhaUncertaintyAwareAnticipationActivities2019} (top-1) & 2019 & \textbf{\ul{74.89}} & \textbf{\ul{58.75}} & \textbf{46.07} & \textbf{\ul{35.71}} & \textbf{67.39} & \textbf{52.37} & \textbf{\ul{46.73}} & \textbf{\ul{36.64}} \\
     & & TempAgg~\cite{senerTemporalAggregateRepresentations2020} & 2020 & 47.20 & 34.60 & 30.50 & 19.10 & 44.80 & 32.70 & 23.50 & 15.30 \\
     & & Zhao\etal~\cite{zhao2020diverse} (avg.) & 2020 & 46.63 & 35.62 & 31.91 & 21.37 & 46.13 & 36.37 & 33.10 & 19.45 \\
     & & Zhao\etal~\cite{zhao2020diverse} (top-1) & 2020 & 51.50 & 38.45 & 36.06 & 27.62 &  50.79 & 47.54 & 37.83 & 29.08 \\
     \midrule
     
    \multirow{7}{*}{Pred. label} & \multirow{7}{*}{Fisher} & RNN~\cite{farhaWhenWillYou2018} & 2018 & 30.06 & 25.43 & 18.74 & 13.49 & 30.77 & 17.19 & 14.79 & 9.77 \\
     & & CNN~\cite{farhaWhenWillYou2018} & 2018 & 21.24 & 19.03 & 15.98 & 9.87 & 29.14 & 20.14 & 17.46 & 10.86 \\
     & & UAAA~\cite{abufarhaUncertaintyAwareAnticipationActivities2019} (avg.) & 2019 & 23.56 & 19.48 & 18.01 & 12.78 & 28.04 & 17.95 & 14.77 & 12.06 \\
     & & UAAA~\cite{abufarhaUncertaintyAwareAnticipationActivities2019} (top-1) & 2019 & \textbf{53.53} & \textbf{42.99} & \textbf{40.50} & \textbf{33.70} & \textbf{56.43} & \textbf{42.82} & \textbf{35.80} & \textbf{30.22} \\
     & & Time-Cond.~\cite{keTimeConditionedActionAnticipation2019} & 2019 & 32.51 & 27.61 & 21.26 & 15.99 & 35.12 & 27.05 & 22.05 & 15.59 \\ 
     & & TempAgg~\cite{senerTemporalAggregateRepresentations2020} & 2020 & 32.70 & 26.30 & 21.90 & 15.60 & 32.30 & 25.50 & 22.70 & 17.10 \\
     & & Piergiovanni\etal~\cite{piergiovanni2020adversarial} & 2020 & 40.40 & 33.70 & 25.40 & 20.90 & 40.70 & 40.10 & 26.40 & 19.20 \\
     
     \midrule
     
     \multirow{13}{*}{Features} & Fisher & TempAgg~\cite{senerTemporalAggregateRepresentations2020} & 2020 & 25.50 & 19.90 & 18.20 & 15.10 & 30.60 & 22.50 & 19.10 & 11.20 \\[0.8mm] 
    
     \arrayrulecolor{lightgray} \hhline{*{1}{~}*{11}{-}}\arrayrulecolor{black} \rule{0pt}{2.6ex}

     & \multirow{12}{*}{I3D} & Cycle Cons.~\cite{abu2020long} & 2020 & 34.76 & 28.41 & 21.82 & 15.25 & 34.39 & 23.70 & 18.95 & 15.89 \\
     & & A-ACT~\cite{gupta2022act} & 2022 & 35.40 & 29.60 & 22.50 & 16.10 & 35.70 & 25.30 & 20.10 & 16.30 \\
     & & FUTR~\cite{gongFutureTransformerLongterm2022} & 2022 & 39.55 & 27.54 & 23.31 & 17.77 & 35.15 & 24.86 & 24.22 & 15.26 \\ 

     & & \rev{DiffAnt}~\cite{zhong2023diffant} & 2023 & 36.13 & 34.00 & 30.46 & 23.77 & 34.09 & 30.14 & 26.34 & 20.23 \\ 
     & & \rev{ActFusion}~\cite{gong2024actfusion} & 2024 & 39.55 & 28.60 & 23.61 & 19.90 & 42.80 & 27.11 & 23.48 & 22.07 \\ 
     & & \rev{GTD}~\cite{zatsarynna2024gated} (avg.) & 2024 & 28.30 & 22.10 & 17.80 & 11.70 & 29.90 & 18.50 & 14.20 & 10.60 \\ 
     & & \rev{GTD}~\cite{zatsarynna2024gated} (top-1) & 2024 & 69.60 & 55.80 & 45.20 & 28.10 & 66.20 & 44.90 & 39.20 & 31.00 \\ 
     & & \rev{ActionLLM}~\cite{wang2025multimodal} & 2025 & 43.67 & 32.80 & 25.73 & 19.51 & 44.95 & 29.06 & 23.15 & 18.37 \\ 
     & & \rev{MANTA}~\cite{zatsarynna2025manta} (avg.) & 2025 & 28.60 & 22.80 & 19.50 & 13.60 & 31.30 & 21.90 & 17.60 & 13.00 \\ 
     & & \rev{MANTA}~\cite{zatsarynna2025manta} (top-1) & 2025 & 68.30 & 51.50 & 41.70 & 31.30 & 71.70 & 53.30 & 43.80 & 31.10 \\ 
     & & \rev{MixANT}~\cite{wasim2025mixant} (avg.) & 2025 & 30.30 & 25.00 & 20.90 & 15.20 & 33.40 & 23.70 & 19.70 & 14.60 \\ 
     & & \rev{MixANT}~\cite{wasim2025mixant} (top-1) & 2025 & \textbf{71.50} & \textbf{56.90} & \textbf{\ul{46.50}} & \textbf{35.00} & \textbf{\ul{72.90}} & \textbf{\ul{54.60}} & \textbf{44.90} & \textbf{32.40} \\ 
     
     \bottomrule
    \end{tabular}
    \label{tab:50salads}
\end{table*}

\end{document}